\documentclass{article}

\usepackage{microtype}
\usepackage{graphicx}
\usepackage{subfigure}
\usepackage{booktabs} 

\usepackage{amsthm}
 
\theoremstyle{definition}

\usepackage{amsmath,amsfonts,bm}

\def\eqref#1{equation~\ref{#1}}

\def\1{\bm{1}}

\DeclareMathAlphabet{\mathsfit}{\encodingdefault}{\sfdefault}{m}{sl}
\SetMathAlphabet{\mathsfit}{bold}{\encodingdefault}{\sfdefault}{bx}{n}

\usepackage{url}
\usepackage{graphicx}
\usepackage{enumitem}

\usepackage{hyperref}

\usepackage[accepted]{icml2020}

\icmltitlerunning{Revisiting Fundamentals of Experience Replay}

\begin{document}

\twocolumn[
\icmltitle{Revisiting Fundamentals of Experience Replay}

\icmlsetsymbol{equal}{*}
\begin{icmlauthorlist}
\icmlauthor{William Fedus}{equal,goo,mila}
\icmlauthor{Prajit Ramachandran}{equal,goo}
\icmlauthor{Rishabh Agarwal}{goo}
\icmlauthor{Yoshua Bengio}{mila,dir}
\icmlauthor{Hugo Larochelle}{goo,cif}
\icmlauthor{Mark Rowland}{deep}
\icmlauthor{Will Dabney}{deep}
\end{icmlauthorlist}

\icmlaffiliation{mila}{MILA, Universit\'e de Montr\'eal}
\icmlaffiliation{goo}{Google Brain}
\icmlaffiliation{deep}{DeepMind}
\icmlaffiliation{dir}{CIFAR Director}
\icmlaffiliation{cif}{CIFAR Fellow}

\icmlcorrespondingauthor{William Fedus}{liamfedus@google.com}

\icmlkeywords{Machine Learning, ICML}

\vskip 0.3in
]



\printAffiliationsAndNotice{\icmlEqualContribution} 

\begin{abstract}
Experience replay is central to off-policy algorithms in deep reinforcement learning (RL), but there remain significant gaps in our understanding.
We therefore present a systematic and extensive analysis of experience replay in Q-learning methods, focusing on two fundamental properties: the replay capacity and the ratio of learning updates to experience collected (\emph{replay ratio}).
Our additive and ablative studies upend conventional wisdom around experience replay --- greater capacity is found to substantially increase the performance of certain algorithms, while leaving others unaffected.
Counterintuitively we show that theoretically ungrounded, uncorrected $n$-step returns are uniquely beneficial while other techniques confer limited benefit for sifting through larger memory.
Separately, by directly controlling the replay ratio we contextualize previous observations in the literature and empirically measure its importance across a variety of deep RL algorithms.
Finally, we conclude by testing a set of hypotheses on the nature of these performance benefits. 
\end{abstract}

\section{Introduction}
Experience replay is the fundamental data-generating mechanism in off-policy deep reinforcement learning \citep{lin1992self}.
It has been shown to improve sample efficiency and stability by storing a fixed number of the most recently collected transitions for training.
However, the \emph{interactions} of experience replay with modern algorithmic components of deep RL agents are still poorly understood.
This lack of understanding impedes progress, as researchers are unable to measure the full impact of algorithmic changes without extensive tuning.
We therefore present a large-scale study to understand the interplay of learning algorithms and data-generating mechanisms in order to inform the design of better algorithms.

Earlier works investigating replay buffers have often focused on individual hyperparameters, such as the capacity of the replay buffer \citep{zhang2017deeperlook},
which has typically been preserved since the seminal work in this field of \citet{mnih2013playing,mnih2015human}.
We begin by identifying that several such hyperparameters, such as buffer capacity and rate of data throughput, are interlinked in the manner that they affect experience replay, modifying both the amount of data available to an agent and the typical age of that data.
This motivates a comprehensive set of experiments to understand the relative effects of modifying each property independently.
We then make the surprising discovery that these effects depend critically on the presence of a particular algorithmic component, $n$-step returns, which has not previously been linked to experience replay.
We conclude by examining several hypotheses to uncover why this link exists.

\section{Background}
We consider a Markov decision process $(\mathcal{S}, \mathcal{A}, P, R, \gamma)$, and denote the sequence of states, actions, and rewards experienced by an agent by $(S_t)_{t \geq 0}$, $(A_t)_{t \geq 0}$, and $(R_t)_{t \geq 0}$, respectively. The central task of reinforcement learning is to find a policy $\pi : \mathcal{S} \rightarrow \Delta_\mathcal{A}$ that maximizes the expected return
\begin{align*}
    Q^\pi(s, a) = \mathbb{E}_\pi \left\lbrack \sum_{t \geq 0} \gamma^t R_t \middle| S_0 = s, A_0 = a \right\rbrack \, ,
\end{align*}
for each initial state-action pair $(s, a) \in \mathcal{S} \times \mathcal{A}$. This problem is well studied, and a wide range of methods based on value iteration, policy iteration, and policy gradients have been built up over the course of decades \citep{Bellman:DynamicProgramming,puterman1994markov,bertsekas196neuro,kaelbling1996reinforcement,szepesvari2010algorithms,sutton2018reinforcement,francois2018introduction}. A prototypical method based on value iteration is Q-learning \citep{watkins1992q}; 
in its most basic form, it maintains an estimate $Q : \mathcal{S} \times \mathcal{A} \rightarrow \mathbb{R}$ of the optimal value function, and given a sequence of transition tuples $(s_t, a_t, r_t, s_{t + 1})_{t \geq 0}$, updates $Q(s_t, a_t)$ towards the target $r_t + \gamma \max_{a \in \mathcal{A}} Q(s_{t + 1}, a)$, for each $t \geq 0$.

\subsection{Deep reinforcement learning} \label{ssec:drl}

In recent years, the field of \emph{deep reinforcement learning} has sought to combine the classical reinforcement learning algorithms mentioned above with modern techniques in machine learning to obtain scalable learning algorithms. 
Deep Q-Networks (DQN) \citep{mnih2015human} combine Q-learning with neural network function approximation and experience replay \citep{lin1992self} to yield a scalable reinforcement learning algorithm that achieves superhuman performance on a range of games in the Arcade Learning Environment \citep{bellemare2013arcade}. Many further approaches have developed since, which like DQN can be understood as comprising three fundamental units:
\begin{enumerate}[leftmargin=0.8cm,topsep=0pt,itemsep=0pt,parsep=0pt,partopsep=0pt,label=(\roman*)]
    \item a function approximation architecture;
    \item a learning algorithm;
    \item a mechanism for generating training data.
\end{enumerate}

A range of innovations in all three areas have been developed since the introduction of the original DQN algorithm. A limited selection of these include architectures based on duelling heads \citep{wang2015dueling} and various forms of recurrence \citep{hausknecht2015deep,kapturowski2018recurrent}, learning algorithms using auxiliary tasks \citep{jaderberg2016reinforcement, fedus2019hyperbolic}, distributional variants of RL \citep{bellemare2017distributional, dabney2018distributional}, and the use of \emph{prioritisation} in sampling from experience replay \cite{schaul2015prioritized}.

A notable agent combining several such innovations is Rainbow \citep{hessel2018rainbow}. An open-source implementation based on this agent is available in Dopamine \citep{castro2018dopamine}, which has four main differences relative to the original DQN agent:
\begin{itemize}[leftmargin=0.5cm,topsep=0pt,itemsep=0pt,parsep=0pt,partopsep=0pt]
    \item \textbf{Prioritized Experience Replay (PER)} \citep{schaul2015prioritized}: A scheme for sampling non-uniformly from the replay buffer that favors transitions with a high temporal-difference (TD) error. In contast, DQN uniformly samples experience from the replay buffer.
    \item \textbf{$n$-step returns}: Rather than training the action-value estimate $Q(s_t, a_t)$ on the basis of the single-step temporal difference error $r_t + \gamma \max_a Q(s_{t+1}, a) - Q(s_t, a_t)$, an $n$-step target $\sum_{k=0}^{n-1} \gamma^k r_{t+k} + \gamma^n \max_a Q(s_{t+n}, a) - Q(s_t, a_t)$ is used, with intermediate actions generated according to a behavior policy $\mu$.
    \item \textbf{Adam optimizer} \citep{kingma2014adam}:  An improved first-order gradient optimizer which normalizes for first and second gradient moments, in contrast to the RMSProp optimizer used by DQN.
    \item \textbf{C51} \citep{bellemare2017distributional}: A distributional RL algorithm that trains an agent to make a series of predictions about the \emph{distribution} of possible future returns, rather than solely estimating the scalar expected return.
\end{itemize}
The Dopamine Rainbow agent differs from that of \citet{hessel2018rainbow} by not including Double DQN updates \citep{van2016deep}, dueling heads \citep{wang2015dueling}, or noisy networks \citep{fortunato2018noisy}. For completeness, we provide a discussion of the details around these algorithmic adjustments in the Appendix.

\subsection{Experience replay}

A critical component of DQN-style algorithms is \emph{experience replay} \citep{lin1992self}. The experience replay is a fixed-size buffer that holds the most recent transitions collected by the policy. It greatly improves the sample efficiency of the algorithm by enabling data to be reused multiple times for training, instead of throwing away data immediately after collection, and also improves the stability of the network during training.

The experience replay is typically implemented as a circular buffer, where the oldest transition in the buffer is removed to make room for a transition that was just collected. Transitions are sampled from the buffer at fixed intervals for use in training. The most basic sampling strategy used is uniform sampling, whereby each transition in the buffer is sampled with equal probability. Other sampling strategies, such as prioritized experience replay \citep{schaul2015prioritized}, can be used instead of uniform sampling. While this is the most common implementation, other variations, such as a distributed experience replay buffer \citep{horgan2018distributed}, can be used.

\citet{mnih2015human} set the experience replay size to hold 1M transitions. This setting is often preserved in works building off DQN \citep{hessel2018rainbow}. In this work, we hold other components of the algorithm fixed and study the effects of modifying various aspects of experience replay.

\subsection{Related work}

While the three principal aspects of DQN-based agents listed in Section~\ref{ssec:drl} have individually received much attention, comparatively little effort has been spent on investigating the interactions between design choices across these areas; notable excamples include the original Rainbow work \citep{hessel2018rainbow}, as well as more recent work focused on $\lambda$-returns and replay \citep{daley2019reconciling}. The principal aim of this work is to improve our understanding of the relationship between data generation mechanisms, in the form of experience replay, and learning algorithms.

\citet{zhang2017deeperlook} study the effects of replay buffer size on performance of agents of varying complexity, noting hyperparameters of the replay buffer such as this are not well understood generally, partly as a result of the complexities of modern learning systems. 
They find that both smaller and larger replay buffers are detrimental to performance on a set three tasks: a gridworld, \textsc{Lunar Lander} \citep{catto2011box2d} and \textsc{Pong} from the Arcade Learning Environment \citep{bellemare2013arcade}. \citet{liu2018effects} also study the effects of replay buffer size and minibatch size on learning performance. 
\citet{fu2019diagnosing} report that agent performance is sensitive to the number of environment steps taken per gradient step, with too small or large a ratio also hindering performance.
\citet{van2019use} vary this ratio in combination with batch sizes to obtain a more sample-efficient version of Rainbow. Beyond direct manipulation of these properties of the buffer, improving and understanding experience replay algorithms remains an active area of research \citep{pan2018organizing,schlegel2019importance,zha2019experience,novati2019remember,sun2020attentive,lee2019sample}.

\section{Disentangling experience replay}

We conduct a detailed study of the ways in which the type of data present in replay affects learning.
In earlier work, \citet{zhang2017deeperlook} studied the effects of increasing the size of the replay buffer of DQN.
We note that in addition to increasing the diversity of samples available to the agent at any moment, a larger replay buffer also typically contains more \emph{off-policy} data, since data from older policies remain in the buffer for longer.
The behavior of agents as we vary the rate at which data enters and leaves the buffer is another factor of variation which is desirable to understand, as this is commonly exploited in distributed agents such as R2D2 \citep{kapturowski2018recurrent}.
Our aim will be to disentangle, as far as is possible, these separate modalities.
To make these ideas precise, we begin by introducing several formal definitions for the properties of replay that we may wish to isolate, in order to get a better understanding behind the interaction of replay and learning.

\subsection{Independent factors of control}

We first disentangle two properties affected when modifying the buffer size.

\newtheorem{definition_nosection}{Definition}[]
\begin{definition_nosection}{}
\textit{The \emph{replay capacity} is the total number of transitions stored in the buffer.}
\end{definition_nosection}
By definition, the replay capacity is increased when the buffer size is increased. A larger replay capacity will typically result in a larger state-action coverage. For example, an $\epsilon$-greedy policy samples actions randomly with probability $\epsilon$, so the total number of random actions in the replay buffer with capacity $N$ will be $\epsilon N$ in expectation.

\begin{definition_nosection}{}
\textit{The \emph{age} of a transition stored in replay is defined to be the number of gradient steps taken by the learner since the transition was generated.
The \emph{age of the oldest policy} represented in a replay buffer is the age of the oldest transition in the buffer.}
\end{definition_nosection}

The buffer size directly affects the age of the oldest policy. This quantity can be loosely viewed as a proxy of the degree of \emph{off-policyness} of transitions in the buffer; intuitively, the older a policy is, the more likely it is to be different from the current policy. However, note that this intuition does not hold for all cases; e.g., if the acting policy cycles through a small set of policies.

Whenever the replay buffer size is increased, both the replay capacity and the age of the oldest policy increase, and the relationship between these two independent factors can be captured by another quantity, the \emph{replay ratio}. 

\begin{definition_nosection}{}
\textit{The \emph{replay ratio} is the number of gradient updates per environment transition.}
\end{definition_nosection}
The replay ratio can be viewed as a measure of the relative frequency the agent is learning on existing data versus acquiring new experience.
The replay ratio stays constant when the buffer size is increased because the replay capacity and the age of the oldest policy both increase.
However, if one of the two factors is independently modulated, the replay ratio will change.
In particular, when the oldest policy is held fixed, increasing the replay capacity requires more transitions per policy, which decreases the replay ratio.
When the replay capacity is held fixed, decreasing the age of the oldest policy requires more transitions per policy, which also decreases the replay ratio.

\begin{figure}[!t]
    \centering
    \includegraphics[width=0.47\textwidth]{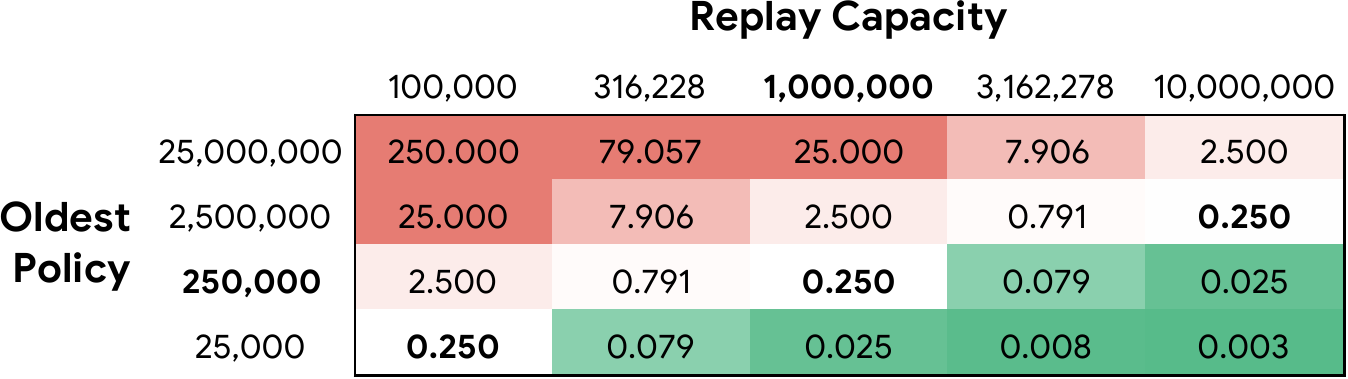}
    \caption{\textbf{Replay ratio varies with replay capacity and the age of the oldest policy.} The replay ratio for controlling different replay capacities (rows) and different ages of the oldest policy (columns). Bold values of \textbf{0.25} are the default replay ratio (one gradient update per four actions) used by \citet{mnih2015human}.}
    \label{fig: replay_ratio}
\end{figure}

In the hyperparameters established in \cite{mnih2015human}, the policy is updated every 4 environment steps collected, resulting in a replay ratio of 0.25.
Therefore, for a replay capacity of 1M transitions, the oldest policy captured in the replay buffer is 250k gradient updates old. 
Figure \ref{fig: replay_ratio} computes the resultant replay ratio when varying either the replay capacity or the age of the oldest policy.
Quantities similar to the replay ratio have also been identified as important hyperparameters in several recent works on deep RL methods \citep{wang2016sample,kapturowski2018recurrent,Hamrick2020Combining,lin2020ranking}. 

\subsection{Experiments}
\label{ssec:grid-experiments}

We conduct experiments on the commonly-used Atari Arcade Learning Environment \citep{bellemare2013arcade} with sticky actions \citep{machado2018revisiting}.
We focus on a subset of 14 games in order to reduce the computational burden of our experiments, as we aim to benchmark a sizeable grid of values for the two factors.
The subset is chosen in a manner meant to reflect the diversity of environments across games (e.g., ensuring we have sparse, hard exploration games such as \textsc{Montezuma's Revenge}).
For each game, we run 3 different random seeds. 
We use Rainbow \citep{hessel2018rainbow} implemented in Dopamine \citep{castro2018dopamine} as our base algorithm in this study due to its strong performance among existing Q-learning agents.

\begin{figure}[!t]
    \centering
    \includegraphics[width=0.47\textwidth]{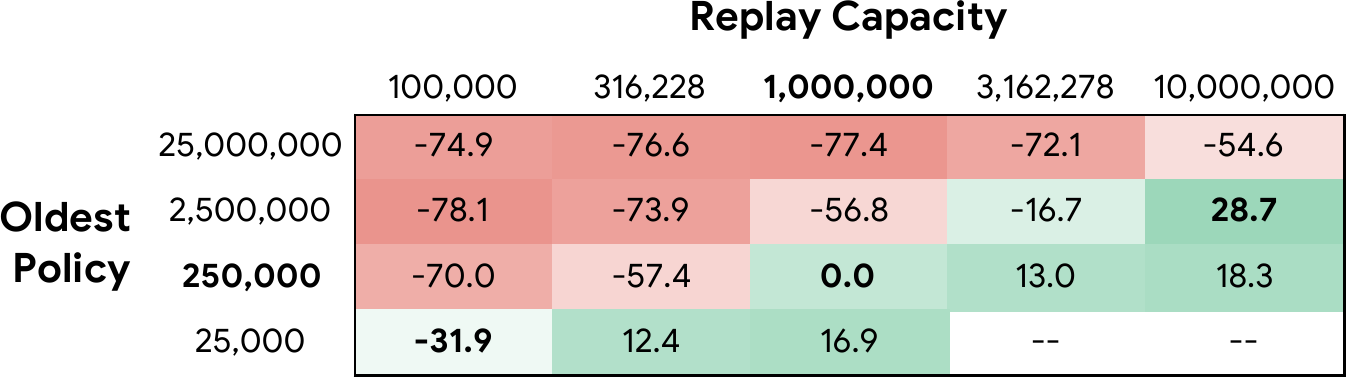}
    \caption{\textbf{Performance consistently improves with increased replay capacity and generally improves with reducing the age of the oldest policy.} Median percentage improvement over the Rainbow baseline when varying the replay capacity and age of the oldest policy in Rainbow on a 14 game subset of Atari. We do not run the two cells in the bottom-right because they are extremely expensive due to the need to collect a large number of transitions per policy.}
    \label{fig: rainbow_grid}
\end{figure}

In these experiments, we fix the total number of gradient updates and the batch size per gradient update to the settings used by Rainbow, meaning that all agents train on the same number of total transitions, although environment frames generated may vary due to controlling for the oldest policy in replay.
Rainbow uses a replay capacity of 1M and an oldest policy of 250k, corresponding to a replay ratio of 0.25.
We assess the cross product of 5 settings of the replay capacity (from 0.1M to 10M) and 4 settings of the oldest policy (from 25k to 25M), but exclude the two settings with the lowest replay ratio as they are computationally impractical due to requiring a large number of transitions per policy.
The replay ratio of each setting is shown in Figure \ref{fig: replay_ratio} and the Rainbow results in Figure \ref{fig: rainbow_grid}. Several trends are apparent.

\paragraph{Increasing replay capacity improves performance. } While fixing the oldest policy, performance improves with higher replay capacity (rows in Figure \ref{fig: rainbow_grid}).
This general trend holds regardless of the particular value of oldest policy, though the magnitude of improvement is dependent on the setting of oldest policy.
It may be that larger replay capacities improves the value function estimates due to having a larger state-action coverage, which can lower the chance of overfitting to a small subset of state-actions.

\begin{figure}[!ht]
  \centering
  \subfigure[\textsc{Asterix}]{\label{fig: asterix_10M}\includegraphics[width=0.85\columnwidth]{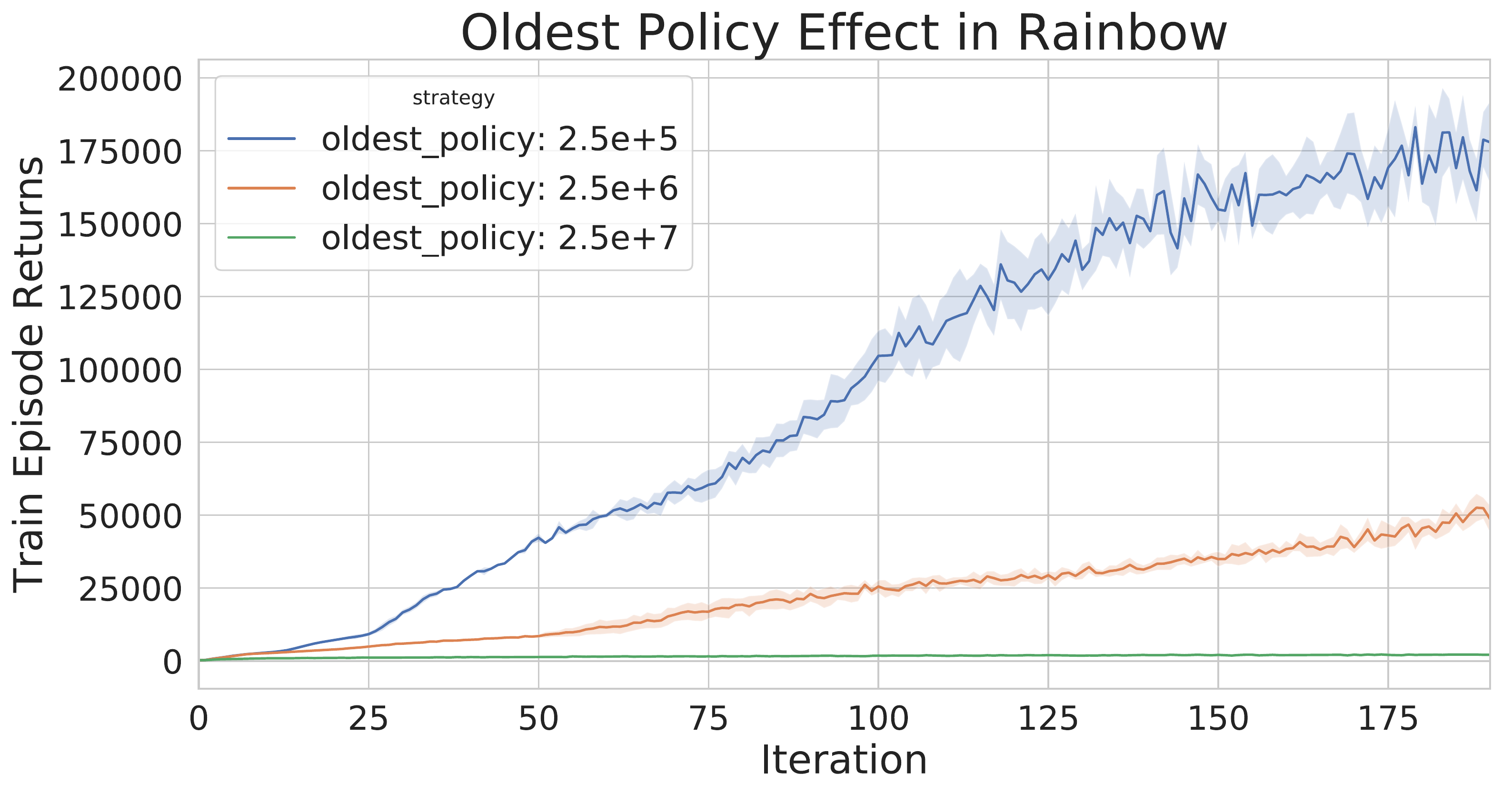}}
  \vfill
  \subfigure[\textsc{Seaquest}]{\label{fig: seaquest_10M}\includegraphics[width=0.85\columnwidth]{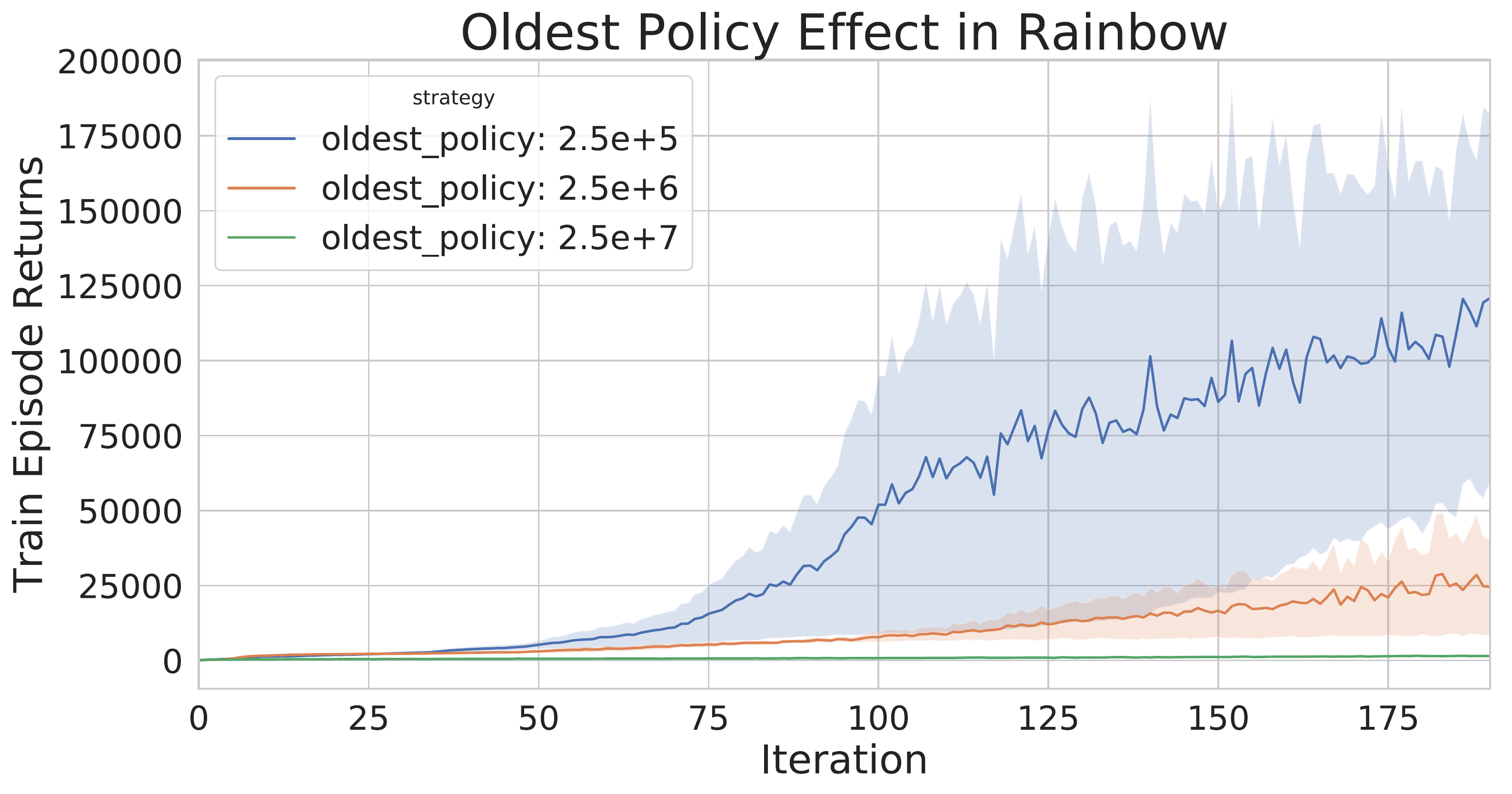}}
  \vfill
  \subfigure[\textsc{PrivateEye}]{\label{fig: privateeye_10M}\includegraphics[width=0.85\columnwidth]{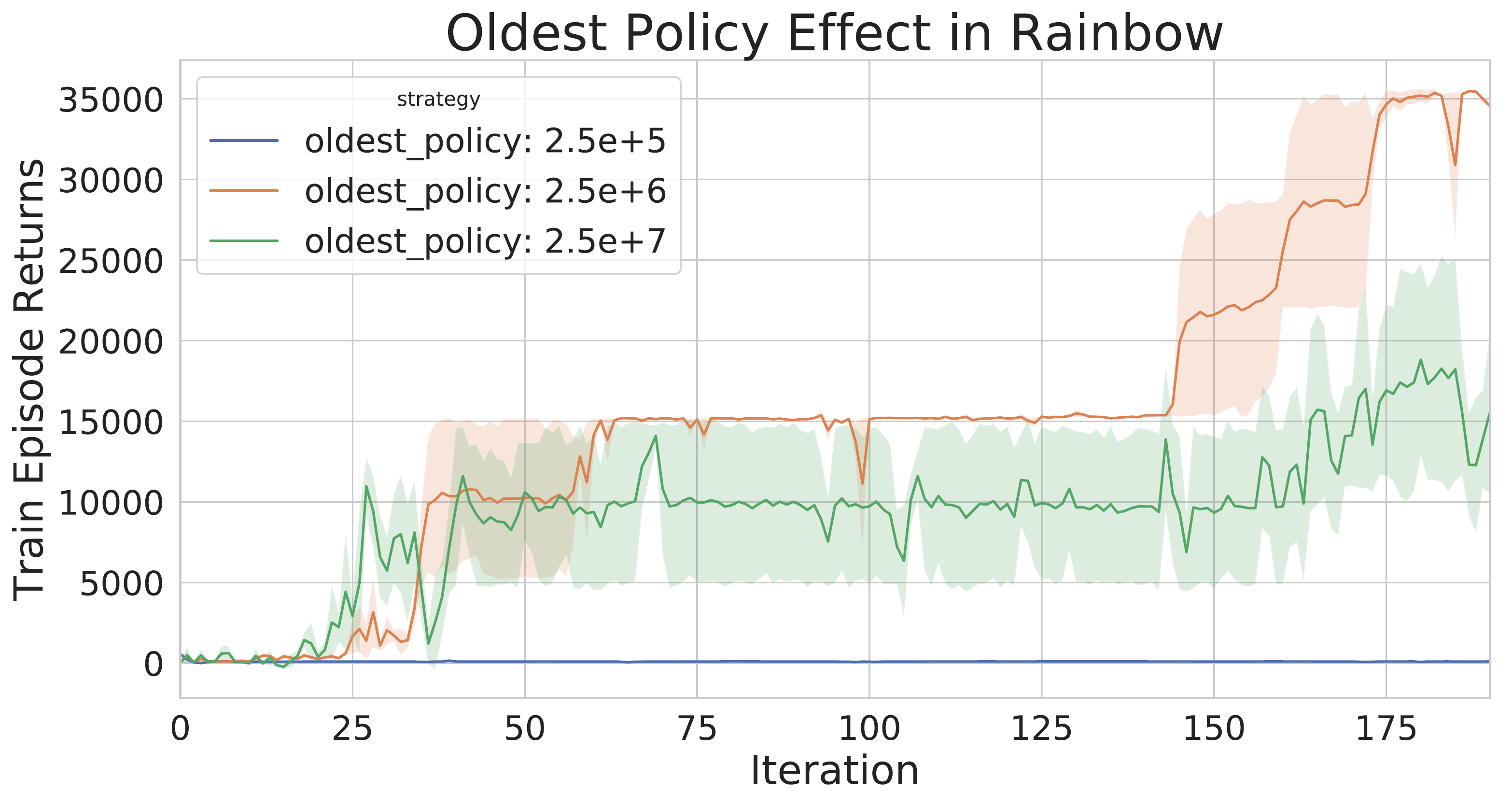}}
  \caption{\textbf{Performance generally improves when trained on data from more recent policies.} Training curves for three games each over a sweep of three oldest policy parameters (2.5e5, 2.5e6 and 2.5e7). Performance generally improves significantly with reduced oldest policies except in sparse-reward games such as \textsc{PrivateEye}.}
  \label{fig: oldest_policy_10M}
  \vspace{-.5em}
\end{figure}

\paragraph{Reducing the oldest policy improves performance.} When fixing the replay capacity, performance tends to improve as the age of the oldest policy decreases (columns in Figure \ref{fig: rainbow_grid}). 
We visualize the training curves for three settings on three games in Figure \ref{fig: oldest_policy_10M}.
Using the heuristic that the age of oldest policy is a proxy for off-policyness, this result suggests that learning from more on-policy data may improve performance.
As the agent improves over the course of training, it spends more time in higher quality (as measured by return) regions of the environment.
Learning to better estimate the returns of high quality regions can lead to further gains.

\begin{figure}[!ht]
    \centering
    \includegraphics[width=0.8\columnwidth]{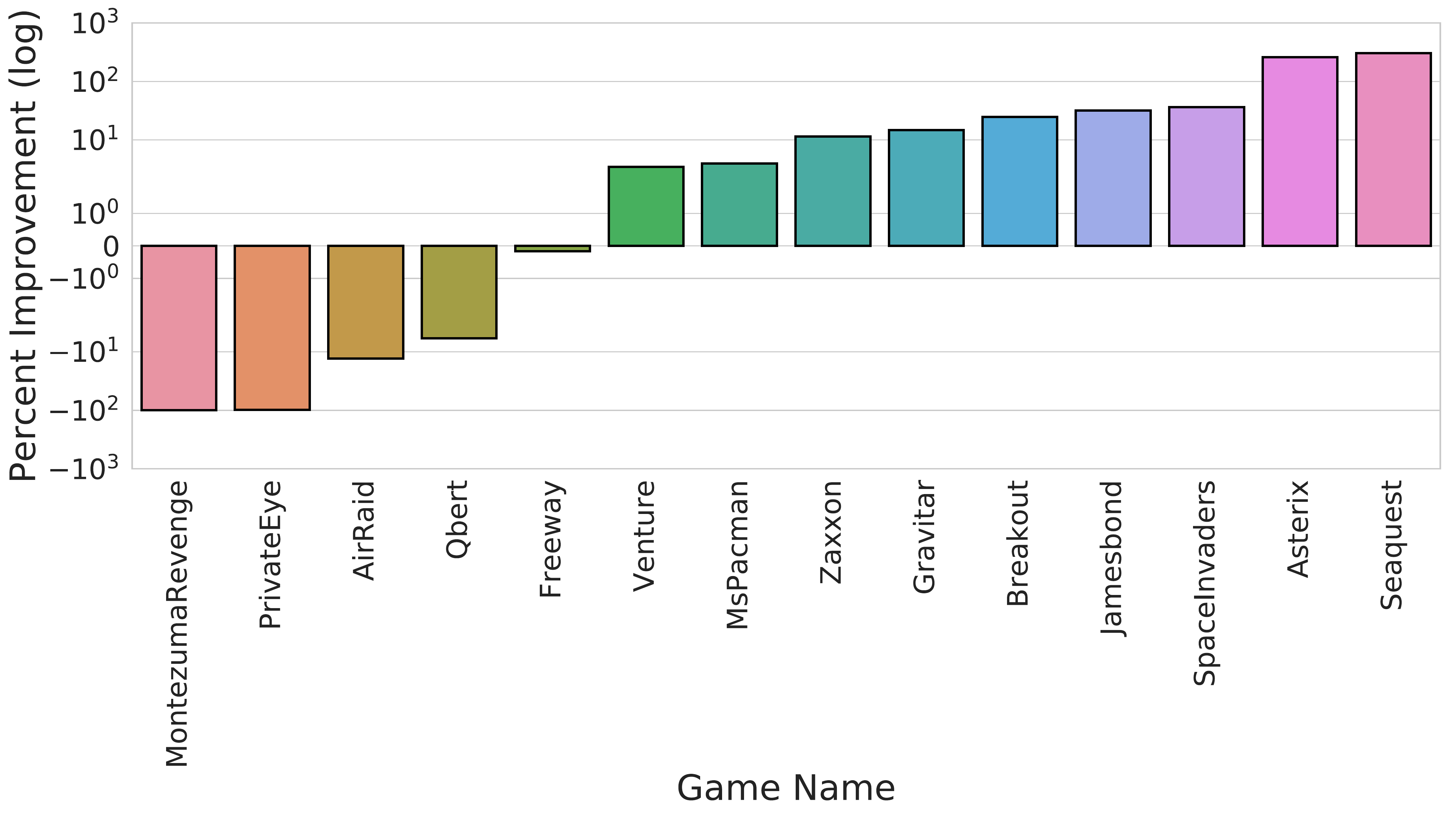}
    \caption{\textbf{Sparse-reward games benefit from data generated by older policies.} Median relative improvement of a Rainbow agent with a 10M replay capacity and 250k oldest policy compared to one with 2.5M oldest policy. Decreasing the age of the oldest policy improves performance on most games. However, performance drops significantly on the two hard exploration games, which bucks the trend that data from newer policies is better.}
    \label{fig: hard-exploration-oldest-policy}
\end{figure}

However, an exception to this trend is seen in the 10M replay capacity setting where the performance drops when moving from an age of 2.5M to 250k. This aberration is explained by a drop in scores of two specific games (see Figure \ref{fig: hard-exploration-oldest-policy}), \textsc{Montezuma's Revenge} and \textsc{PrivateEye}, which are considered sparse-reward hard-exploration environments \citep{bellemare2016unifying}.
Agents that only sample from the newest policies do not seem to be able to find the sparse reward (see Figure \ref{fig: oldest_policy_10M}(c)).

\paragraph{Increasing buffer size with a fixed replay ratio has varying improvements.} When the replay ratio is fixed while the buffer size is modified, there is an interplay between the improvements caused by increasing the replay capacity and the deterioration caused by having older policies in the buffer. The magnitude of both effects depends on the particular settings of these quantities. Generally, as the age of the oldest policy increases, the benefits from increasing the replay capacity are not as large. 

\subsection{Generalizing to other agents}
The experiments in the previous subsection are conducted using the Dopamine Rainbow algorithm but we now test whether experience replay behaves similarly in other Q-learning variants.
In particular, we test if increases in the replay capacity improve performance with the original DQN algorithm \citep{mnih2015human}.

We maintain the default Dopamine hyperparameters \citep{castro2018dopamine} specifically tuned for DQN and increase the replay capacity from 1M to 10M.
We consider two experimental variants: fixing either the replay ratio or the oldest policy. 
Fixing the replay ratio corresponds to the standard setting of increasing the buffer size hyperparameter.
Fixing the oldest policy requires adjusting the replay ratio so that the replay buffer always contains policies within a certain number of gradient updates.
The results are presented in Table \ref{table: dqn-1-to-10}.

\begin{table}[!ht]
    \caption{\textbf{DQN does not improve with larger replay capacities, unlike Rainbow.} Relative improvements of increasing replay capacity from 1M to 10M for DQN and Rainbow. Relative improvements are computed with respect to performance of the corresponding agent with a 1M replay capacity. Either the replay ratio or the oldest policy is held fixed when increasing the replay capacity.}
    \label{table: dqn-1-to-10}
    \begin{tabular}{c|cc}
    \toprule
    \textbf{Agent}  & \textbf{Fixed replay ratio} & \textbf{Fixed oldest policy} \\ \midrule
    DQN      & +0.1\%                & -0.4\%                 \\
    Rainbow & +28.7\%               & +18.3\%                \\ \bottomrule
    \end{tabular}
\end{table}

Surprisingly, providing a DQN agent with an order of magnitude larger memory confers no benefit regardless of whether the replay ratio or the oldest policy is held fixed.
These results stand in contrast to the dynamics of the Rainbow agent which demonstrates consistent improvements with increased replay capacity.
We also note the fixed replay ratio result disagrees with the conclusion in \citet{zhang2017deeperlook} that larger replay capacity is \emph{detrimental} -- we instead observe no material performance change.

This result calls into question which differences between these two value-based algorithms are driving the distinct responses to an increased replay buffer size. 
In the next section, we perform a large scale study to determine which algorithmic components enable Rainbow to take advantage of a larger replay capacity. 

\section{What components enable improving with a larger replay capacity?}
As described in Section \ref{ssec:drl}, the Dopamine Rainbow agent is a DQN agent with four additional components: prioritized experience replay \citep{schaul2015prioritized}, $n$-step returns, Adam \citep{kingma2014adam}, and C51 \citep{bellemare2017distributional}.
We therefore seek to attribute the performance difference under larger replay capacities to one or more of these four components.
To do so, we study agents built from a variety of subsets of components, and measure whether these variant agents improve when increasing the replay capacity.
In these studies we specifically measure the \emph{relative} improvement upon increasing the replay capacity, not which variant achieves the highest absolute return.

\subsection{Additive and ablative experiments}
\begin{figure}[!ht]
    \centering
    \includegraphics[width=0.4\textwidth]{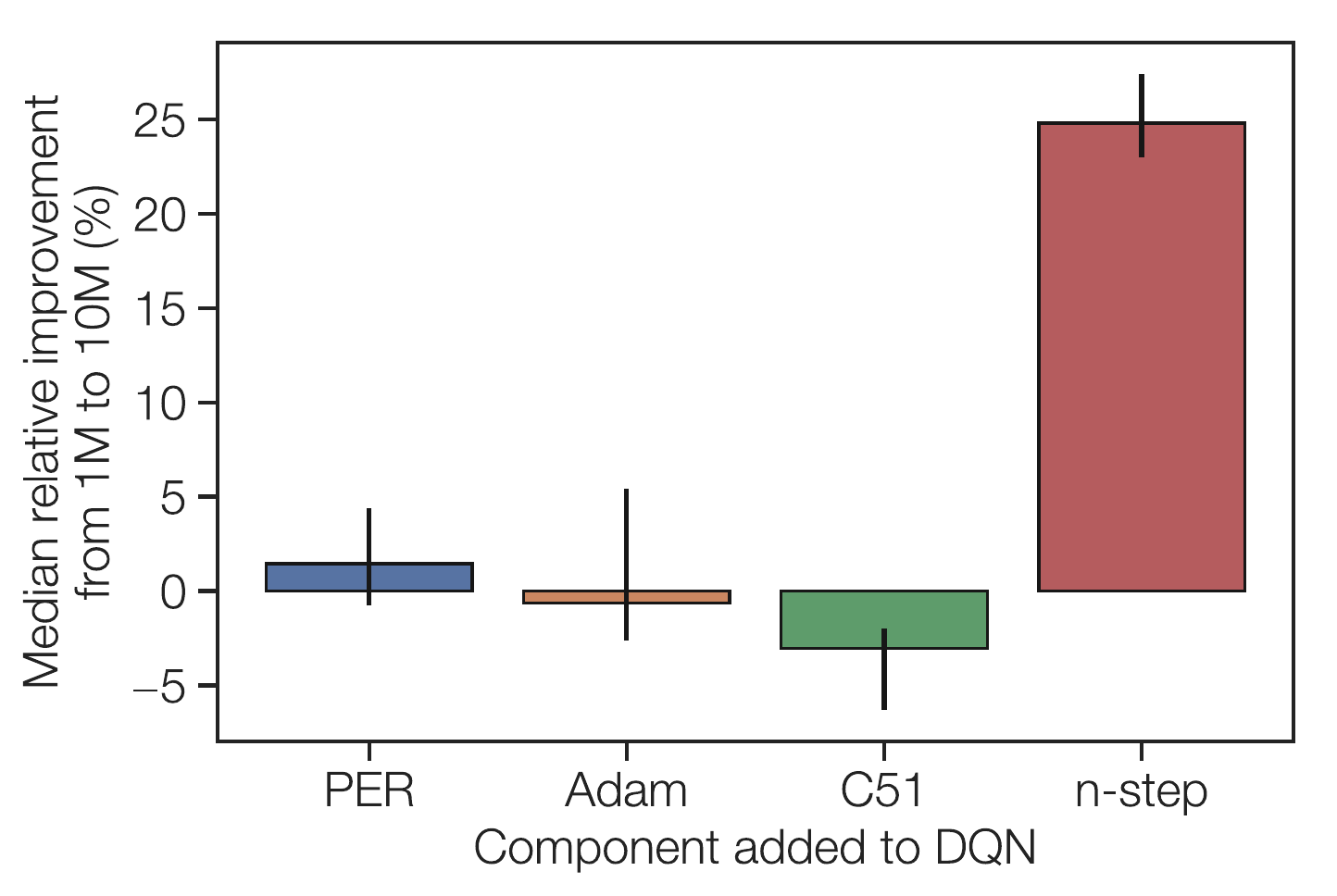}
    \caption{\textbf{Adding $n$-step to DQN enables improvements with larger replay capacities.} Median relative improvement of DQN additive variants when increasing replay capacity from 1M to 10M. Bars represent 50\% percentile improvement and the lower and upper bound of the error line is denoted by 25\% and 75\% percentiles, respectively.}
    \label{fig:dqn-additions}
\end{figure}

We begin with an \emph{additive} study where we add a single component of Rainbow to the DQN algorithm, resulting in four agent variations.
We independently compute the relative performance difference when increasing the replay capacity from 1M to 10M for each variant.
When increasing the replay capacity, we fix the replay ratio, and revisit the case of fixing the oldest policy later in the section.
We evaluate across a set of 20 games that is a superset of the 14 games used in the previous section. 
The results are shown in Figure \ref{fig:dqn-additions}.

\begin{figure}[!ht]
    \centering
    \includegraphics[width=0.4\textwidth]{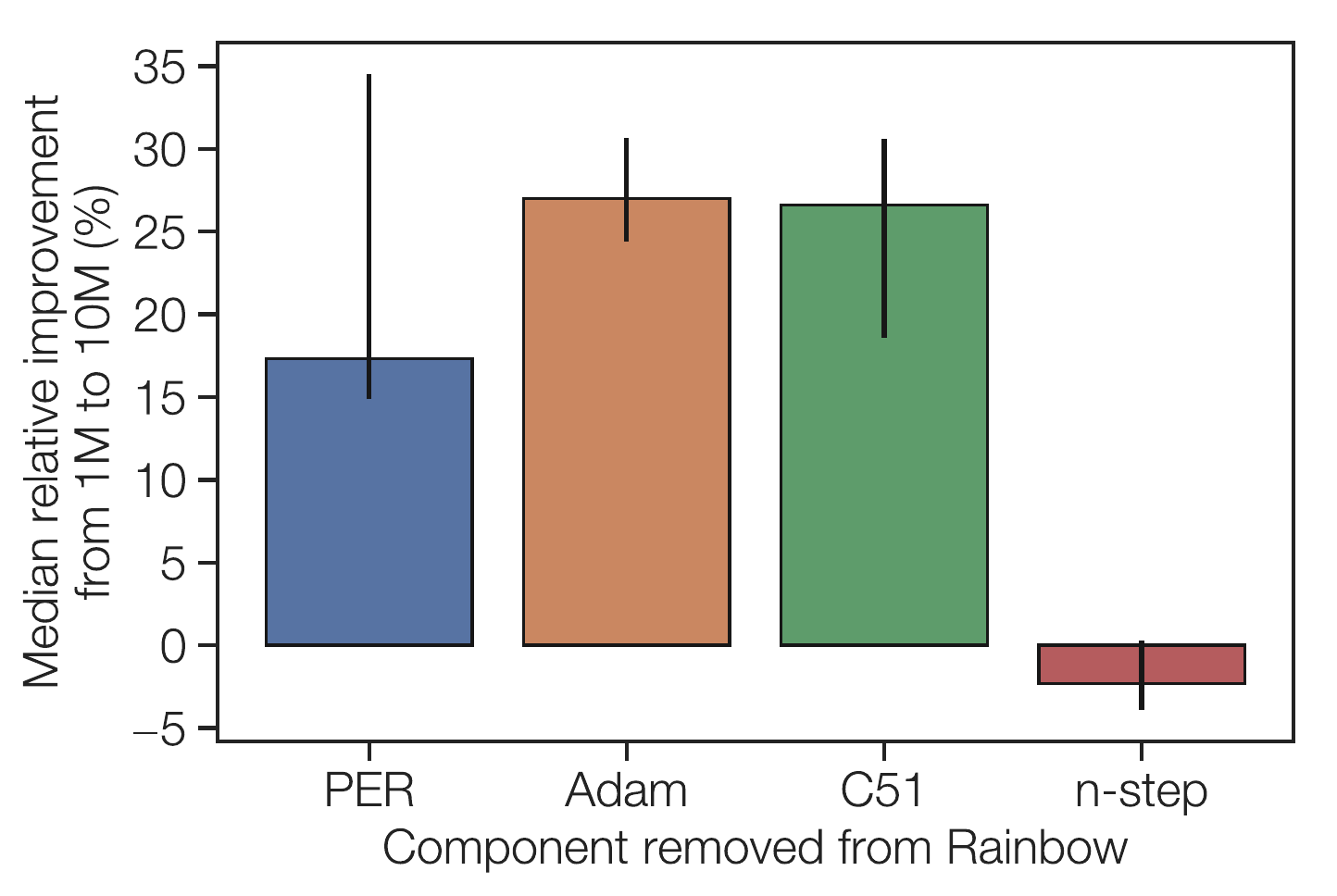}
    \caption{\textbf{Removing $n$-step from Rainbow prevents improvements with larger replay capacities.} Median relative improvement of Rainbow ablative variants when increasing replay capacity from 1M to 10M.  Bars represent 50\% percentile improvement and the lower and upper bound of the error line is denoted by 25\% and 75\% percentiles, respectively.}
    \label{fig:rainbow-ablations}
\end{figure}

The only additive variant that materially improves with larger replay capacity is the DQN agent with $n$-step returns.
From this we hypothesize that $n$-step returns uniquely confer a benefit with larger replay capacity.
As a test of this hypothesis, removing the $n$-step returns from the Rainbow agent should inhibit this ablative variant from improving with a larger replay capacity.
Furthermore, for $n$-step returns to be the \emph{sole} influential component, the ablative versions of the other three components (PER, Adam, C51) must still show improvements with a larger replay capacity.
We present the result of this ablative Rainbow experiment in Figure \ref{fig:rainbow-ablations}.

As predicted, a Rainbow agent stripped of $n$-step returns does not benefit with larger replay capacity, while the Rainbow agents stripped of other components still improve.
These results suggest that $n$-step returns are \emph{uniquely} important in determining whether a Q-learning algorithm can improve with a larger replay capacity.
Another surprising finding is that prioritized experience replay does not significantly affect the performance of agents with larger memories;
intuitively, one might expect prioritized experience replay to be useful in selecting relevant experience for the learner as the replay buffer size grows.
Further detail at the per-game level is provided in Appendix \ref{appendix: additions_ablations}.

\begin{figure}[!ht]
    \centering
    \includegraphics[width=0.4\textwidth]{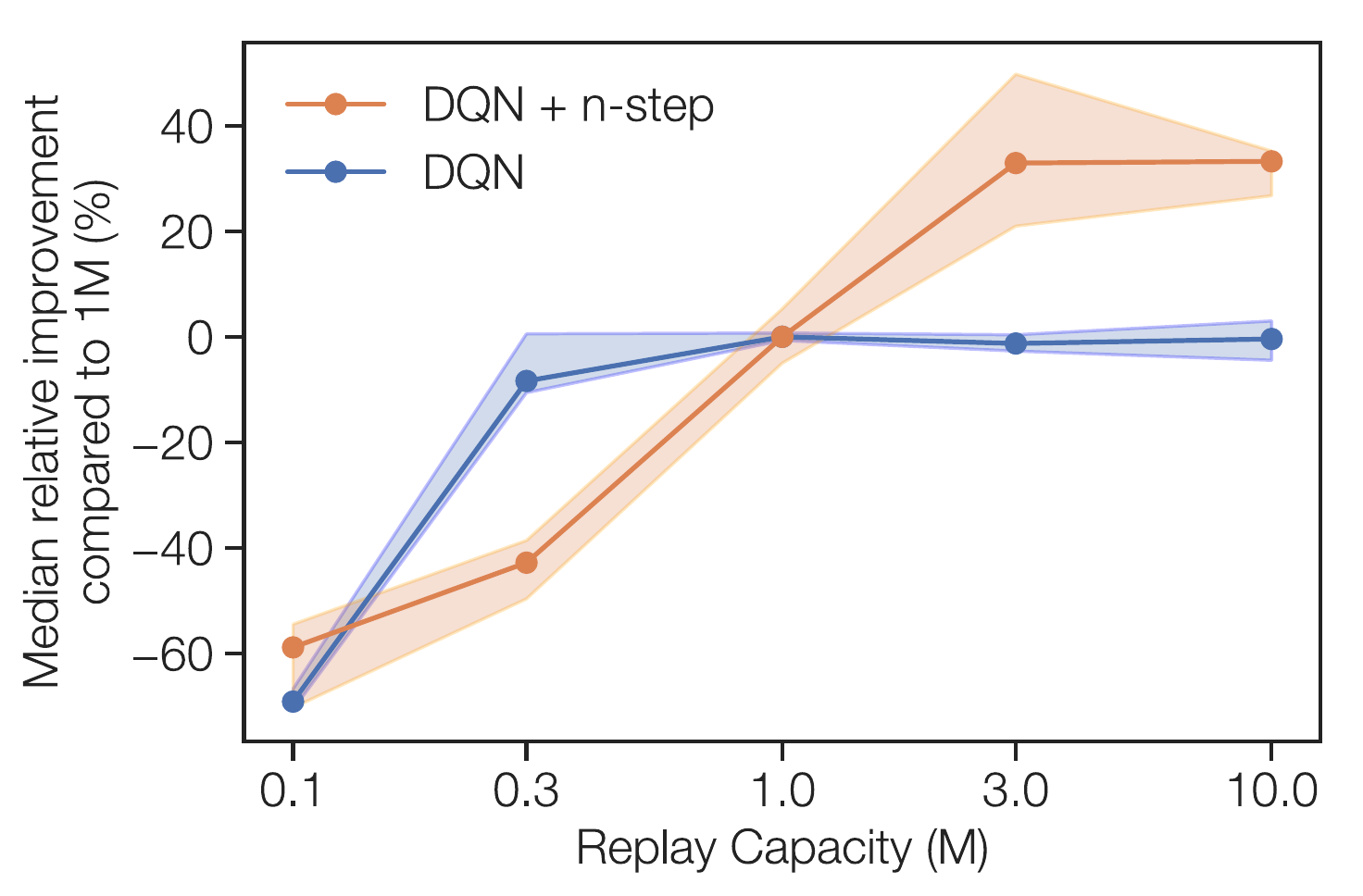}
    \caption{\textbf{DQN + $n$-step improve with larger replay capacity if oldest policy is fixed.} $25^{\text{th}}$, $50^{\text{th}}$, and $75^{\text{th}}$ percentile relative improvement of DQN and DQN + $n$-step when increasing replay capacity while the oldest policy is fixed at 250k.}
    \label{fig:dqn-fixed-oldest-policy}
\end{figure}

As one final control, we check that DQN with $n$-step returns still improves with larger replay capacity if the \emph{oldest policy} is held fixed, rather than the replay ratio being held fixed.
Figure \ref{fig:dqn-fixed-oldest-policy} shows that the DQN + $n$-step algorithm is able to consistently improve when increasing the replay capacity from the highly tuned default of 1M while the standard DQN does not. When given less data, DQN with $n$-step can perform worse, an observation that we revisit in Section \ref{ssec:variance-reduction}.

Taken together, these results suggest that $n$-step is a critical factor for taking advantage of larger replay sizes.
This is unexpected.
Uncorrected $n$-step returns are not theoretically justified for off-policy data because they do not correct for differences between the behavior and target policies, but they are still used due to convenience and their empirical benefits.
The experiments show that extending this theoretically unprincipled approach into a regime where issues may be further exacerbated\footnote{Larger buffers under a fixed replay ratio will contain data from older policies which potentially increases the discrepancy between the old behavior and current agent.} is essential for performance.

\subsection{$n$-step for massive replay capacities}\label{sec:batch}

These results suggest that $n$-step returns should be used when increasing replay capacity.
However, our previous results only consider replay capacities up to 10M, which is a fraction of the 200M total transitions collected over the entire course of training.
It may be the case that $n$-step is no longer as beneficial, or even harmful, as the replay capacity increases.
This degradation may happen because when fixing the replay ratio, which is the most common setting used in practice, the age of the oldest policy will increase alongside the replay capacity.
As demonstrated in Section \ref{ssec:grid-experiments}, the exact settings of each factor controls the magnitude of degradation caused by an increase in oldest policy and the magnitude of improvement caused by an increase in replay capacity.
Furthermore, the uncorrected nature of $n$-step returns may hurt performance in regimes of high off-policyness.

\begin{figure}[!ht]
    \centering
    \includegraphics[width=0.4\textwidth]{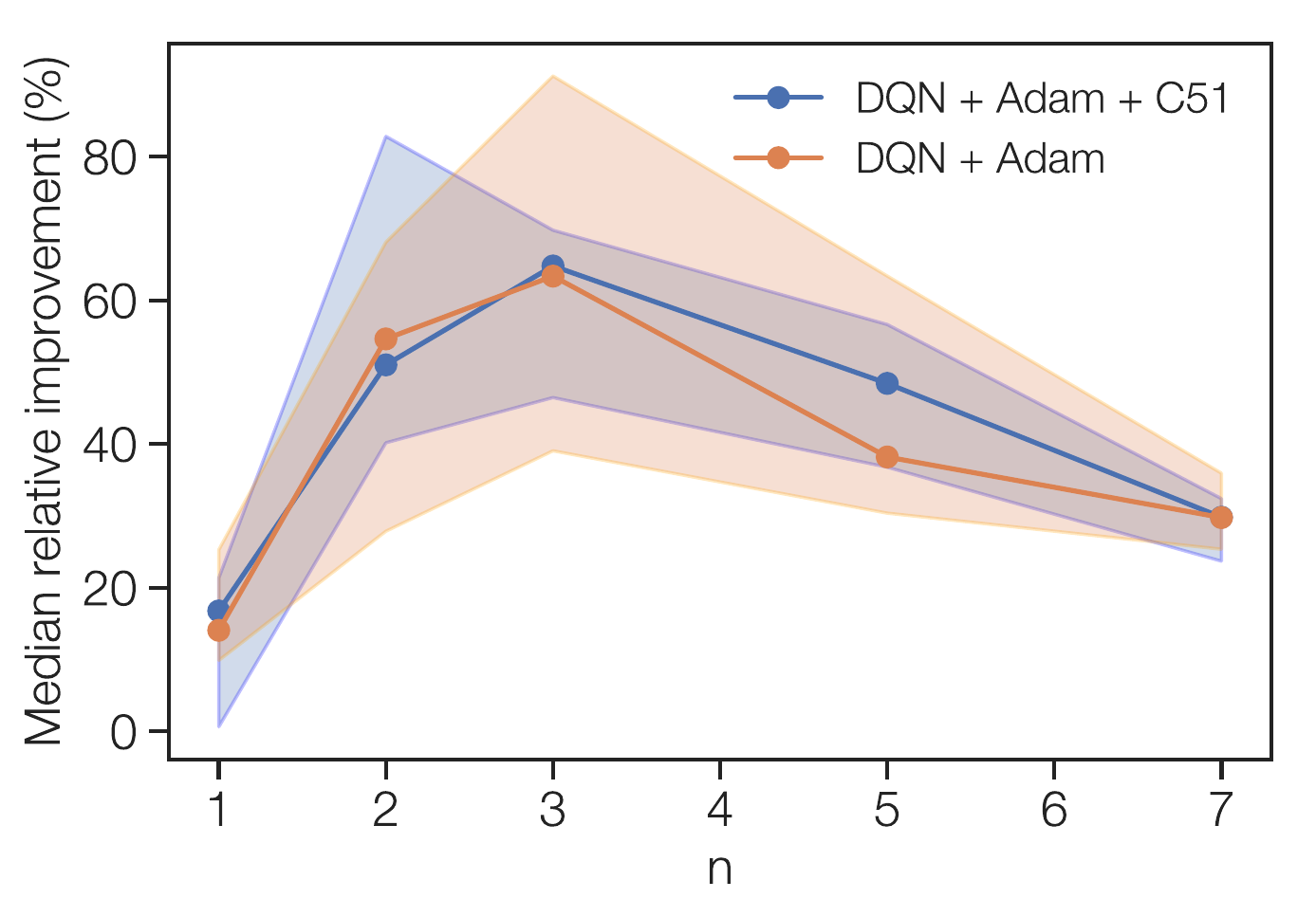}
    \caption{\textbf{Agents improve with $n$-step in the offline batch RL setting.} $25^{\text{th}}$, $50^{\text{th}}$, and $75^{\text{th}}$ percentile relative improvement of each offline agent over the DQN used to gather the training data. Using $n > 1$ improves performance.}
    \label{fig:batch-rl}
\end{figure}

Therefore to test the limits of the hypothesis that $n$-step returns are useful in large capacity regimes with high levels of off-policyness, we turn to the logical extreme --- offline deep reinforcement learning \citep{agarwal2019striving}.
In offline reinforcement learning, a learner is trained only using data collected from another agent.
All data from the original agent is preserved unlike the typical case in online reinforcement learning where older experience is evicted from the replay buffer.
This also represents a worst-case scenario with respect to off-policyness because the learner cannot interact with the environment to correct its estimates.
We use the settings of \citet{agarwal2019striving} where for each game in the same subset of games used in previous experiments, a DQN agent collects a data set of 200M frames which are used to train another agent.
We train two variants of DQN with $n$-step returns, and compare each setting of $n$ against the online DQN agent used to generate the data.
The results are presented in Figure \ref{fig:batch-rl}.

Even in this challenging task, using $n > 1$ consistently improves performance for both agents.
The shape of the curve when varying $n$ depends on the particular agent that is used, but setting $n = 3$, which is the value used in all previous experiments, performs well.
These results further validate the hypothesis that $n$-step is beneficial when increasing the replay capacity.

\section{Why is $n$-step the enabling factor?}
In the previous section, we showed empirically that $n$-step returns modulates whether DQN can take advantage of larger replay capacities. In this section, we attempt to uncover the mechanism that links these two seemingly unrelated components together. In the hypotheses that we evaluate, we find that one plays a partial role in the linkage between $n$-step and replay capacity.

\subsection{Deadening the deadly triad}

Function approximation of Q-values, bootstrapping, and off-policy learning have been identified as the \emph{deadly triad} \citep{sutton2018reinforcement, van2018deep} of properties that, when combined, can negatively affect learning or even cause divergence. \citet{van2018deep} suggest that $n$-step returns work well because they make the magnitude of the bootstrap smaller, making divergence less likely. Recall that the $n$-step target is $\sum_{k=0}^{n-1} \gamma^k r_{t+k} + \gamma^n \max_a Q(s_{t+n}, a)$ where $\gamma \in [0, 1)$ is the discount factor and $\gamma^n$ is the contraction factor. The smaller the contraction factor, the less impact the bootstrap $\max_a Q(s_{t+n}, a)$ has on the target. 

When the replay capacity is increased while keeping the replay ratio fixed, the transitions in the buffer come from older policies, which may increase the off-policyness of the data and, according to the deadly triad, destabilize training. Thus, one may hypothesize that the supposed stability offered by $n$-step is required to counter the increased off-policyness produced by a larger replay capacity.

We test this hypothesis by applying a standard 1-step update to DQN with the same contractive factor as an $n$-step update: $r_{t} + \gamma^n \max_a Q(s_{t+1}, a)$; this is equivalent to simply reducing the discount factor, although we note that it also changes the fixed point of the algorithm. If the contractive factor is the key enabler, using DQN with the modified update should be able to improve with increased replay capacity. However, we find empirically that there is no improvement with an increased replay capacity when using the smaller contractive factor in a 1-step update.  Furthermore, even if the oldest policy is fixed, which should control off-policyness, DQN does not improve with a larger capacity (see Figure \ref{fig:dqn-fixed-oldest-policy}). These results suggests that the hypothesis that the stability improvements of $n$-step that arise from the lower contraction rate do not explain the importance of $n$-step in taking advantage of larger replay capacities.

\subsection{Variance reduction}
\label{ssec:variance-reduction}

One can view $n$-step returns as interpolating between estimating Monte Carlo (MC) targets, $\sum_{k=0}^{T} \gamma^k r_{t+k}$, and single-step temporal difference (TD) targets, $r_{t} + \gamma \max_a Q(s_{t+1}, a)$.
It balances between the low bias but high variance of MC targets, and the low variance but high bias of single-step TD targets.
The variance of MC targets comes from stochasticity of rewards and environmental dynamics, whereas the bias of single-step TD targets comes from using an imperfect bootstrap to estimate future returns. 

An increase in replay capacity might provide a means of mitigating the additional variance of $n$-step returns, relative to single-step TD targets. The increased variance of the $n$-step target increases the learning algorithm's sensitivity to changes in the replay buffer data. Whilst an increased replay capacity will not affect the variance of the learner's target ascribable to minibatch sampling from a fixed replay buffer, it will affect the diversity of transitions which the buffer contains. Thus, in scenarios where the data-generating policy is rapidly changing, for example, a small replay buffer may undergo wild shifts in the type of data in contains, which may have a particularly pronounced effect on higher variance $n$-step methods. In contrast, a larger buffer may moderate the effects of fluctuations in the data-generating policy.

This brief analysis provides a testable hypothesis: in an environment with less variance in returns, the gains from increasing the replay capacity should be reduced.
The variance of returns in the Atari domain can be reduced by turning off \emph{sticky actions}. 
Sticky actions \citep{machado2018revisiting} cause the previously taken action to be repeated with some probability -- increasing the stochasticity of the transition dynamics -- which in turn increases the variance of returns.

\begin{figure}[!ht]
    \centering
    \includegraphics[width=0.4\textwidth]{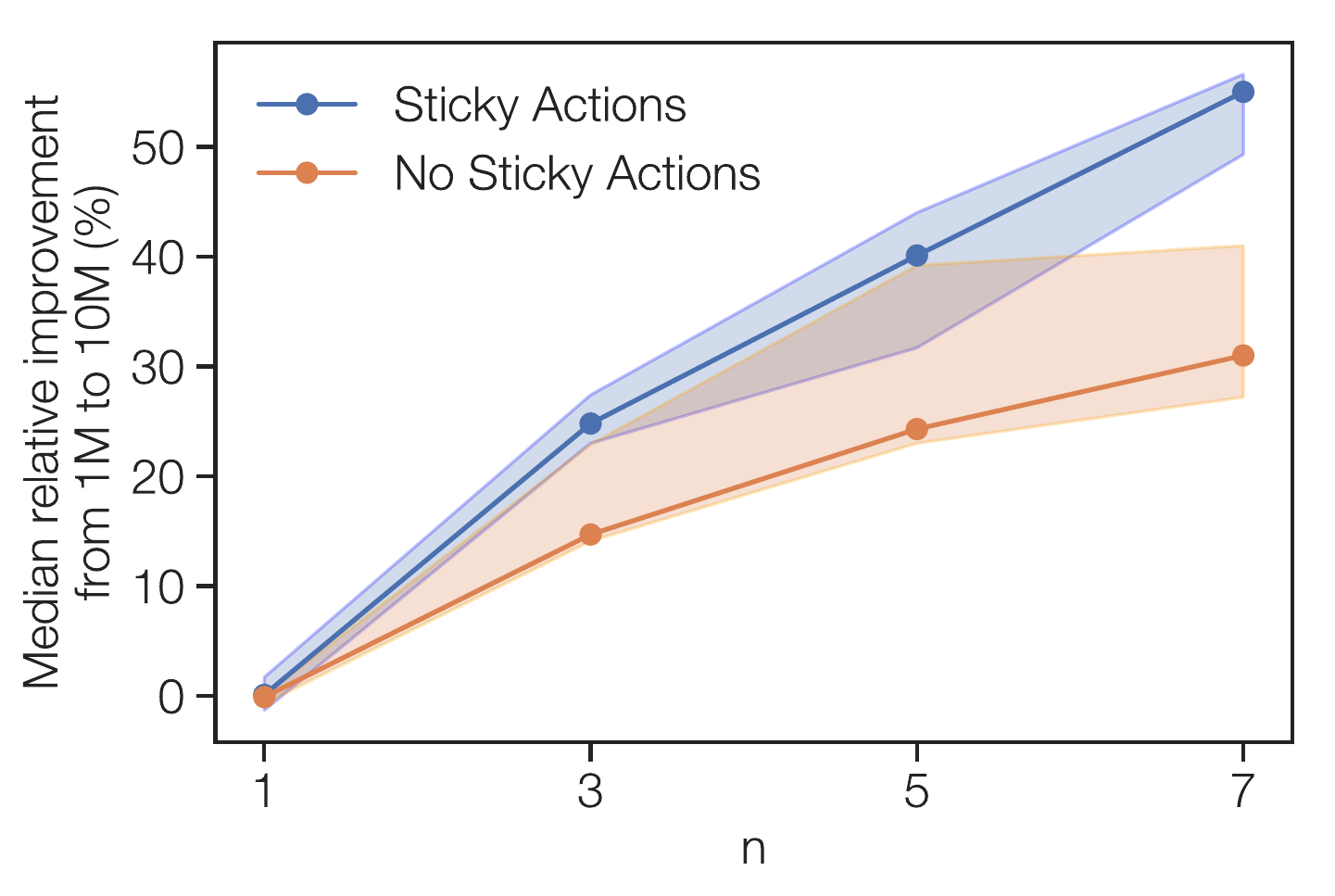}
    \vspace{-10px}
    \caption{\textbf{Turning off sticky actions reduces gains from increasing replay capacity.} $25^{\text{th}}$, $50^{\text{th}}$, and $75^{\text{th}}$ percentile relative improvements from increasing the replay capacity from 1M to 10M for DQN agents with $n$-step when toggling sticky actions. Larger $n$ benefit more from increased replay capacity, but agents trained in an environment without sticky actions benefit less than agents trained in an environment with sticky actions.}
    \label{fig:sticky-actions}
\end{figure}

We test this hypothesis by running 1-,3-,5- and 7-step versions of DQN on the ALE with and without sticky actions, and report results in Figure \ref{fig:sticky-actions}.
As predicted by the hypothesis, the relative improvements in the absence of sticky actions are consistently less than than the relative improvements with sticky actions present.
Furthermore, the difference of improvements between sticky actions and no sticky actions increases with $n$, which is predicted by the hypothesis given that variance also increases with $n$.
However, even when removing stochasticity, using $n$-step returns still shows improvements with increased capacity, indicating that whilst there is some evidence for the hypothesis presented here, it can only play a partial part in explaining the effectiveness of $n$-step updates and their ability to make use of larger buffers.

\subsection{Further multi-step and off-policy methods}

Our investigation has focused specifically on the effects of $n$-step returns, as one of the key aspects of the Rainbow agent. These findings naturally open further questions as to the interaction between experience replay and more general classes of return estimators based on multi-step, off-policy data, such as variants of Q($\lambda$) \citep{watkins1989learning,PENG1994226,sutton2014new,harutyunyan2016q}, TreeBackup \citep{Precup2000EligibilityTF} and Retrace \citep{munos2016safe}, which we believe will be interesting topics for future work.

\section{Discussion}

We have conducted an in-depth study of how replay affects performance in value-based deep reinforcement learning agents. The summary of our contributions are:
\begin{enumerate}[leftmargin=0.8cm,topsep=0pt,itemsep=0pt,parsep=0pt,partopsep=0pt,label=(\alph*)]
\item Disentangling the effects of \emph{replay capacity} and \emph{oldest policy}, finding that increasing replay capacity and decreasing the age of the oldest policy improves performance;
\item Discovering that $n$-step returns are uniquely critical for taking advantage of an increased replay capacity;
\item Benchmarking $n$-step returns in the massive replay capacity regime, and finding that it still provides gains despite the substantial off-policyness of the data;
\item Investigating the connection between $n$-step returns and experience replay, and finding that increasing the replay capacity can help mitigate the variance of $n$-step targets, which partially explains the improved performance.
\end{enumerate}

Taking a step back, this can be interpreted as an investigation into how two of the principal aspects of deep RL agents described in Section~\ref{ssec:drl}, namely learning algorithms and data generating mechanisms, interact with one another. These two aspects are inextricably linked; the data on which an algorithm is trained clearly affects what is learned, and correspondingly what is learned affects how the agent interacts with the environment, and thus what data is generated. We highlight several fundamental properties of the data generating distribution: (a) Degree of on-policyness (how close is the data-generating distribution to the current policy being evaluated?); (b) State-space coverage; (c) Correlation between transitions; (d) Cardinality of distribution support.

Practically, these aspects may be difficult to control independently, and the typical algorithmic adjustments we can make affect several of these simultaneously; two examples of such adjustments are the replay capacity and replay ratio investigated in this paper. We emphasize that these practically controllable aspects of an agent may also have differing effects on the data distribution itself depending on the precise architecture of the agent; for example, in a distributed agent such as R2D2 \citep{kapturowski2018recurrent}, decreasing the replay ratio by increasing the number of actors will lead to changes in both (a) and (c) above, whilst in a single-actor agent such as DQN, changing the replay ratio by altering the number of environment steps per gradient step will also change (b).

These issues highlight the \emph{entanglement} that exists between these different properties of the data-generating mechanism at the level of practical algorithmic adjustments, and motivates further study into how these properties can be disentangled. This direction of research is particularly important with regard to obtaining agents which can effortlessly scale with increased availability of data.
More broadly, this work opens up many questions about the interaction of replay and other agent components, the importance of $n$-step returns in deep RL, and off-policy learning, which we expect to be interesting subjects for future work.

\section*{Acknowledgements}
We'd like to thank Carles Gelada and Jacob Buckman for many lively discussions trying to understand early empirical results.
In addition, we thank Dale Schuurmans for theoretical insights and Sylvain Gelly for advice on conducting a hard-nosed scientific study.
We had several helpful discussions with the Google Brain RL team, in particular, Dibya Ghosh and Marlos Machado.
Finally, we would like to thanks Georg Ostrovski for extensive comments on an earlier draft of this work.

\bibliography{references}
\bibliographystyle{icml2019}

\clearpage
\onecolumn
\appendix
\section*{\centering APPENDICES: Revisiting Fundamentals of Experience Replay}

\section{Experimental details}

\subsection{The Dopamine Rainbow agent}

Our empirical investigations in this paper are based on the Dopamine Rainbow agent \citep{castro2018dopamine}. This is an open source implementation of the original agent \citep{hessel2018rainbow}, but makes several simplifying design choices. The original agent augments DQN through the use of (a) a distributional learning objective, (b) multi-step returns, (c) the Adam optimizer, (d) prioritized replay, (e) double Q-learning, (f) duelling architecture, and (g) noisy networks for exploration. The Dopamine Rainbow agent uses just the first four of these adjustments, which were identified as the most important aspects of the agent in the original analysis of \citet{hessel2018rainbow}.

\subsection{Atari 2600 games used}

A 14 game subset was used for the grid measuring the effects of varying replay capacity and oldest policy. A 20 game subset, which is comprised of the 14 games used for the grid with 6 additional games, was used for all other experiments.

\paragraph{14 game subset:} \textsc{Air Raid}, \textsc{Asterix}, \textsc{Breakout}, \textsc{Freeway}, \textsc{Gravitar}, \textsc{James Bond}, \textsc{Montezuma's Revenge}, \textsc{Ms. Pacman}, \textsc{Private Eye}, \textsc{Q*bert}, \textsc{Seaquest}, \textsc{Space Invaders}, \textsc{Venture}, \textsc{Zaxxon}.

\paragraph{20 game subset:} The 14 games above in addition to: \textsc{Asteroids}, \textsc{Bowling}, \textsc{Demon Attack}, \textsc{Pong}, \textsc{Wizard of Wor}, \textsc{Yars' Revenge}.

\section{Additive and ablative studies}\label{appendix: additions_ablations}
\subsection{DQN additions}
\begin{figure}[b]
    \centering
    \null
    \hfill
    \subfigure[DQN + 3-step provides a median performance change of +24.8\%.]{\label{fig: dqn_3_step}\includegraphics[width=0.4\textwidth]{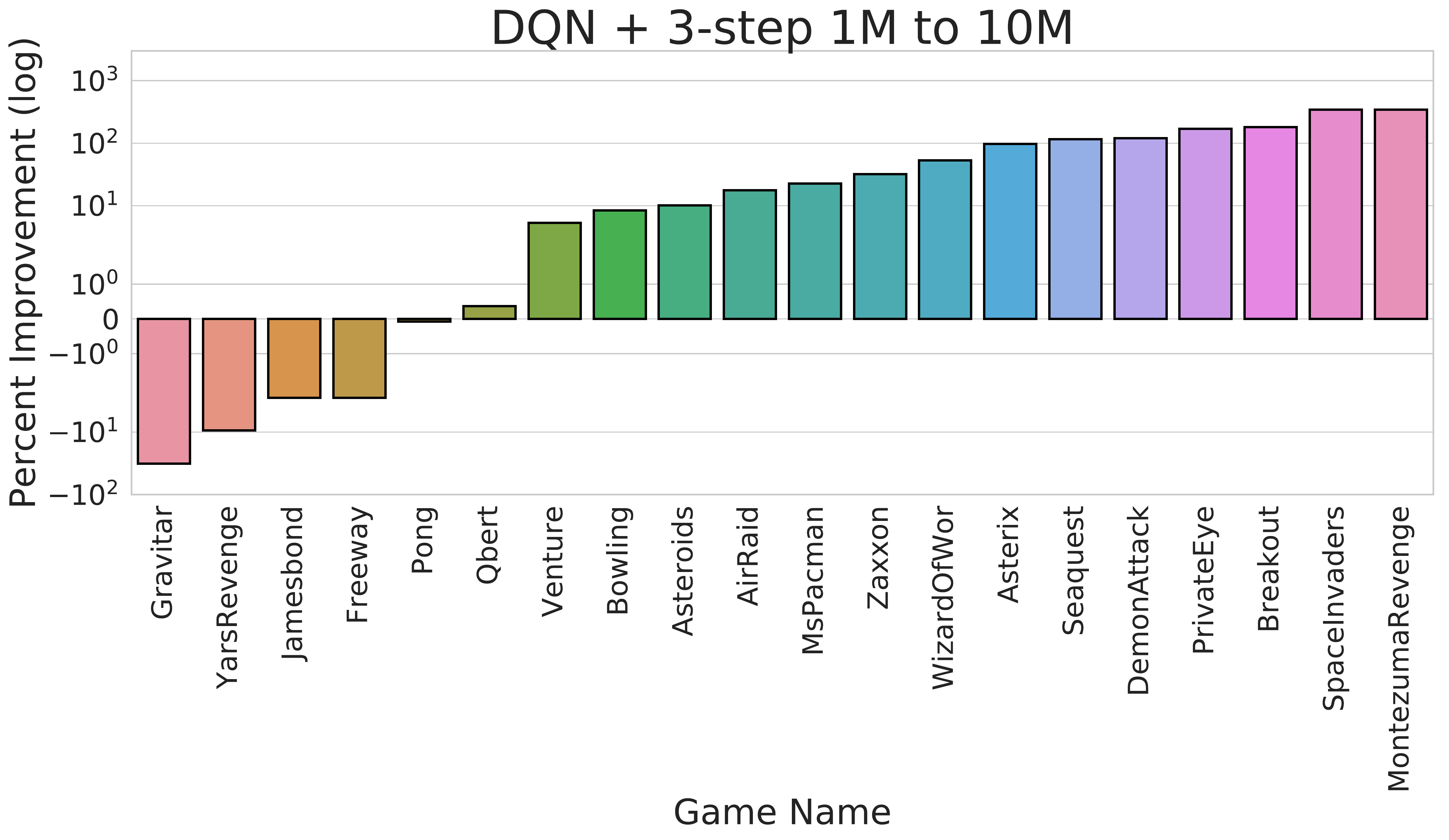}}
    \hfill
    \subfigure[DQN + PER provides a median performance change of +1.5\%.]{\label{fig: dqn_prioritized}    \includegraphics[width=0.4\textwidth]{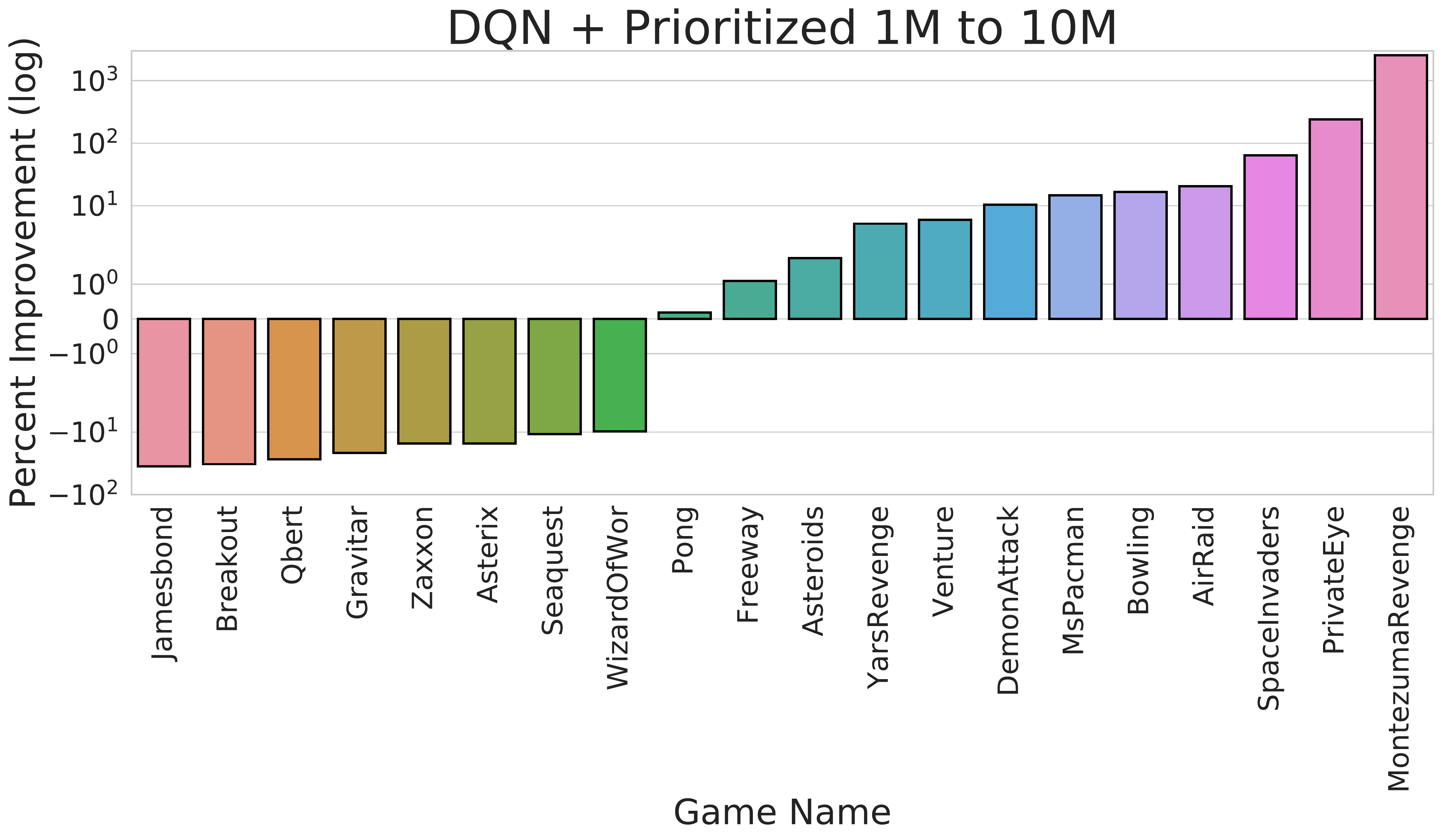}}
    \hfill
    \null
    
    \null
    \hfill
    \subfigure[DQN + Adam provides a median performance change of -0.6\%.]{\label{fig: dqn_adam}\includegraphics[width=0.4\textwidth]{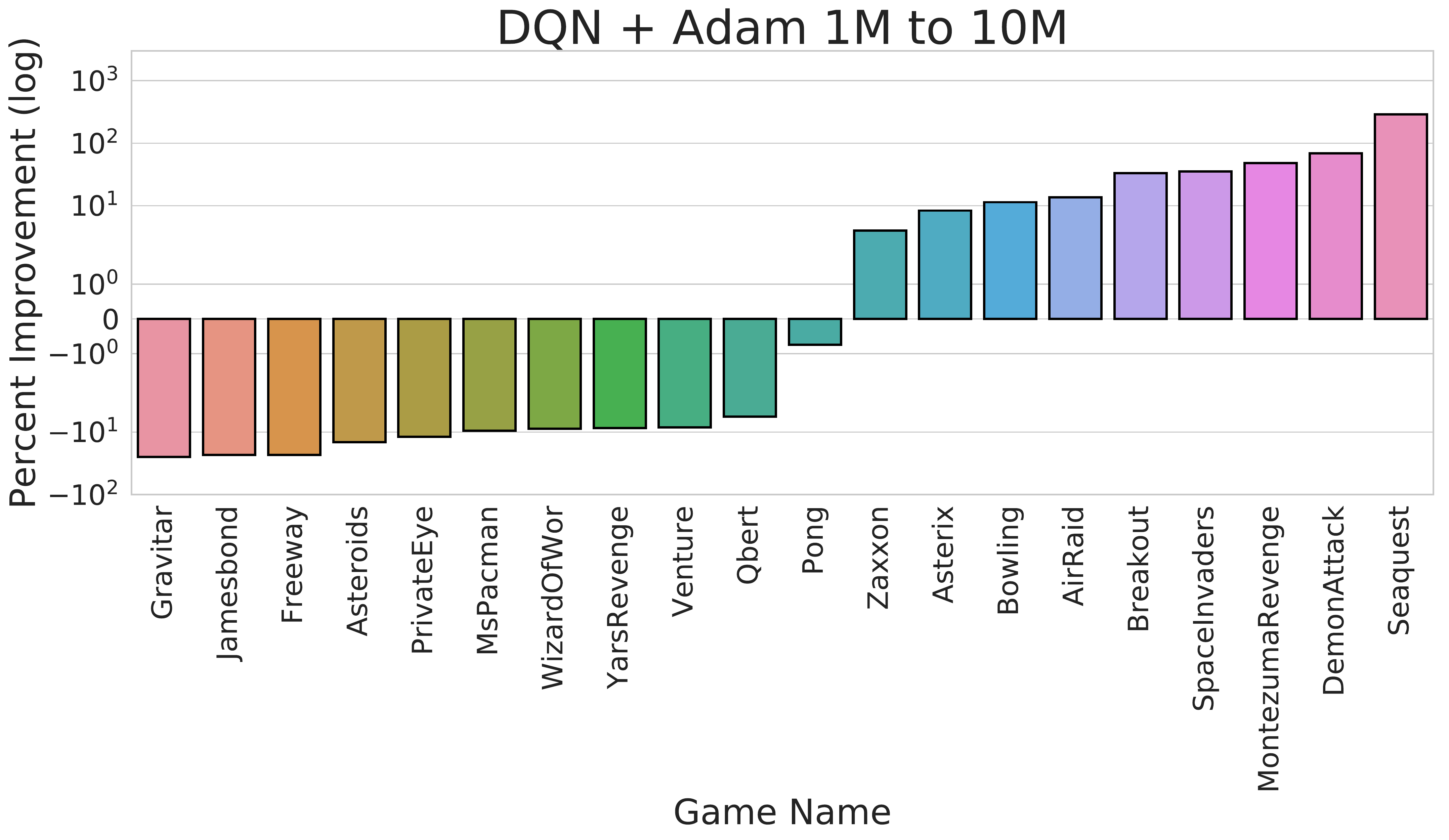}}
    \hfill
    \subfigure[DQN + C51 provides a median performance change of -3.0\%.]{\label{fig: dqn_c51}    \includegraphics[width=0.4\textwidth]{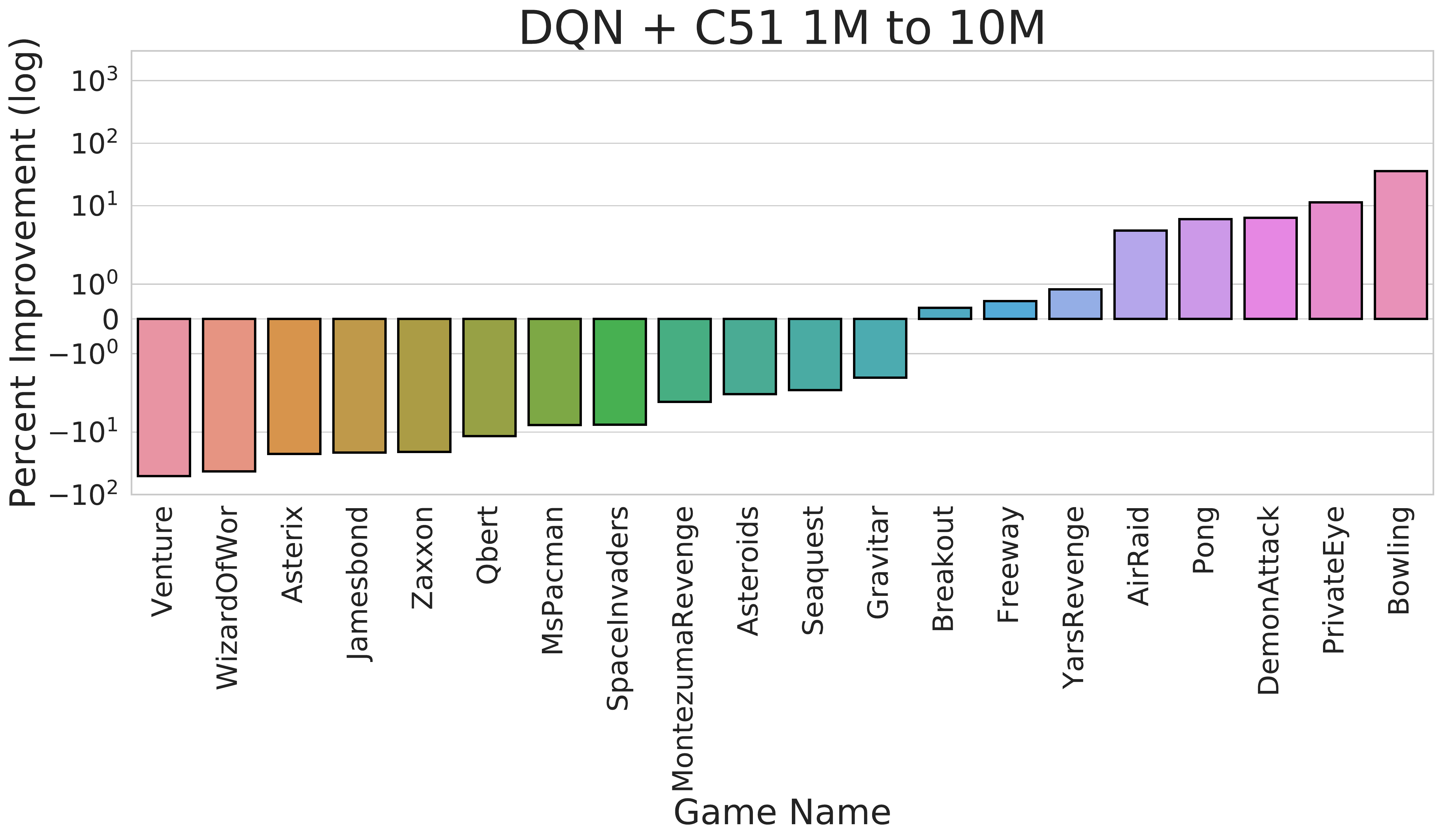}}
    \hfill
    \null

    \caption{\textbf{Only DQN with $n$-step improves with increased capacity.} DQN with an additional component results at a per-game level, measuring performance changes when increasing replay capacity from 1M to 10M.}
    \label{fig: dqn_augmented}
\end{figure}

We provide game-level granularity on the performance of each supplemented DQN agent in Figure \ref{fig: dqn_augmented}. 

\subsection{Rainbow ablations}
\begin{figure}
    \centering
    \null
    \hfill
    \subfigure[The performance difference for a Rainbow agent without n-step returns when increasing the replay buffer size from 1M to 10M. We find that the resultant agent does not benefit from larger replay buffer sizes, reporting a median performance decrease of 2.3\%.  This implies the importance of n-step returns.]{\label{fig: rainbow_n_step_10M_log}\includegraphics[width=0.4\textwidth]{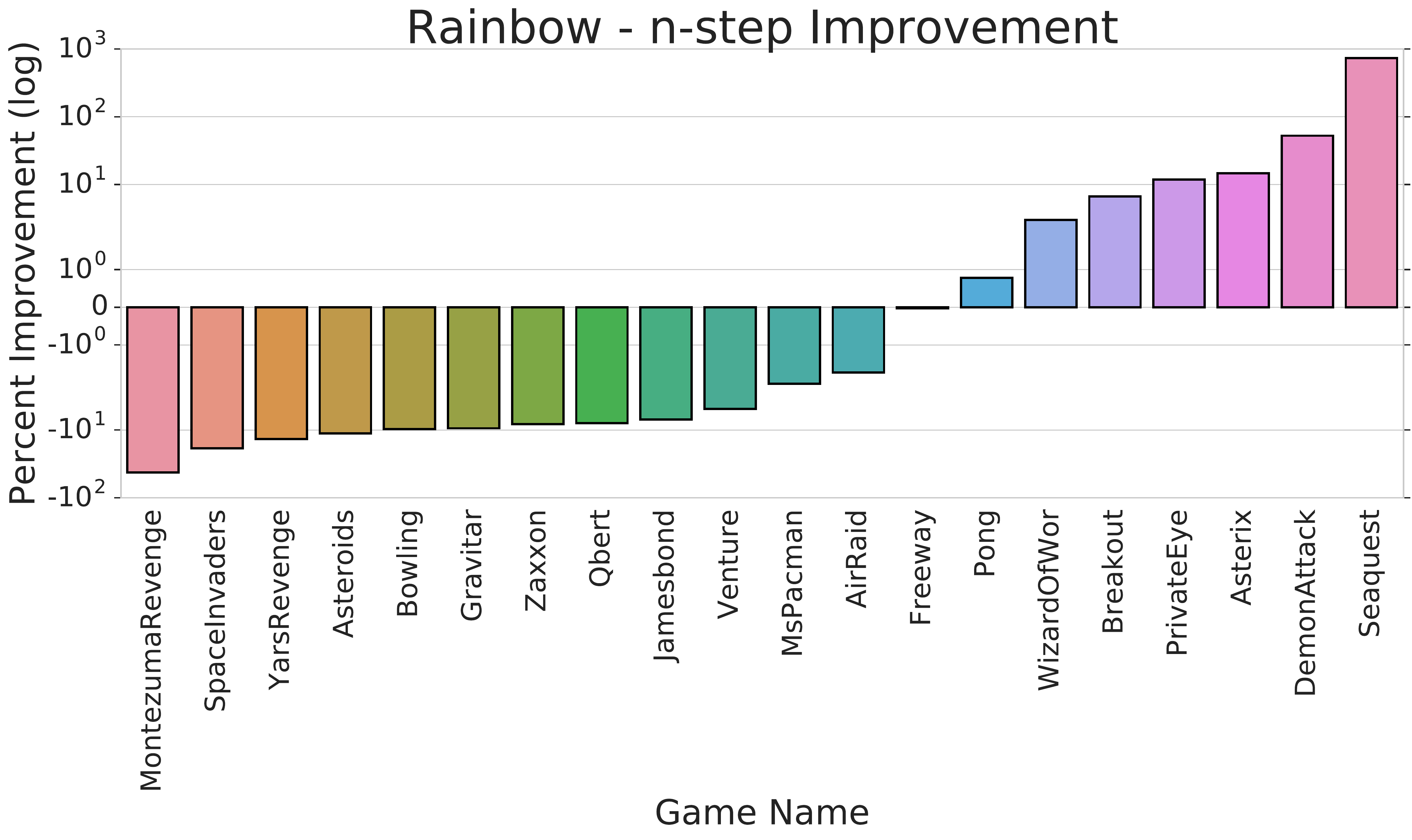}}
    \hfill
    \subfigure[The performance difference for a Rainbow agent without prioritized experience replay when increasing the replay buffer size from 1M to 10M. Even without prioritization, the algorithm still benefits +17.3\%.]{\label{fig: rainbow_prioritized_10M_log}    \includegraphics[width=0.4\textwidth]{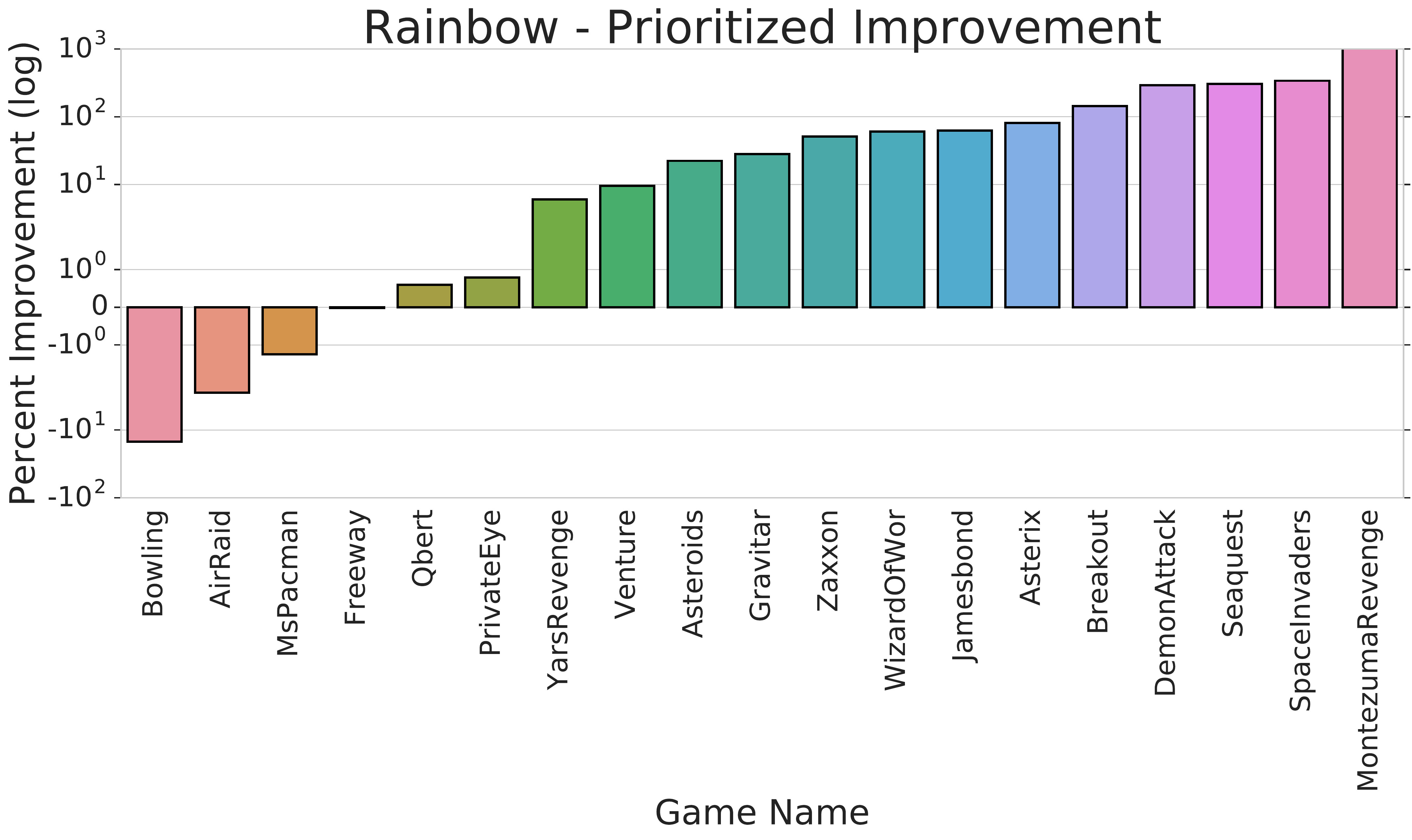}}
    \hfill
    \null
    
    \null
    \hfill
    \subfigure[A Rainbow agent without Adam optimizer has a median performance increase of +27.0\% when the replay buffer size is increased from 1M to 10M.]{\label{fig: rainbow_adam_10M_log}\includegraphics[width=0.4\textwidth]{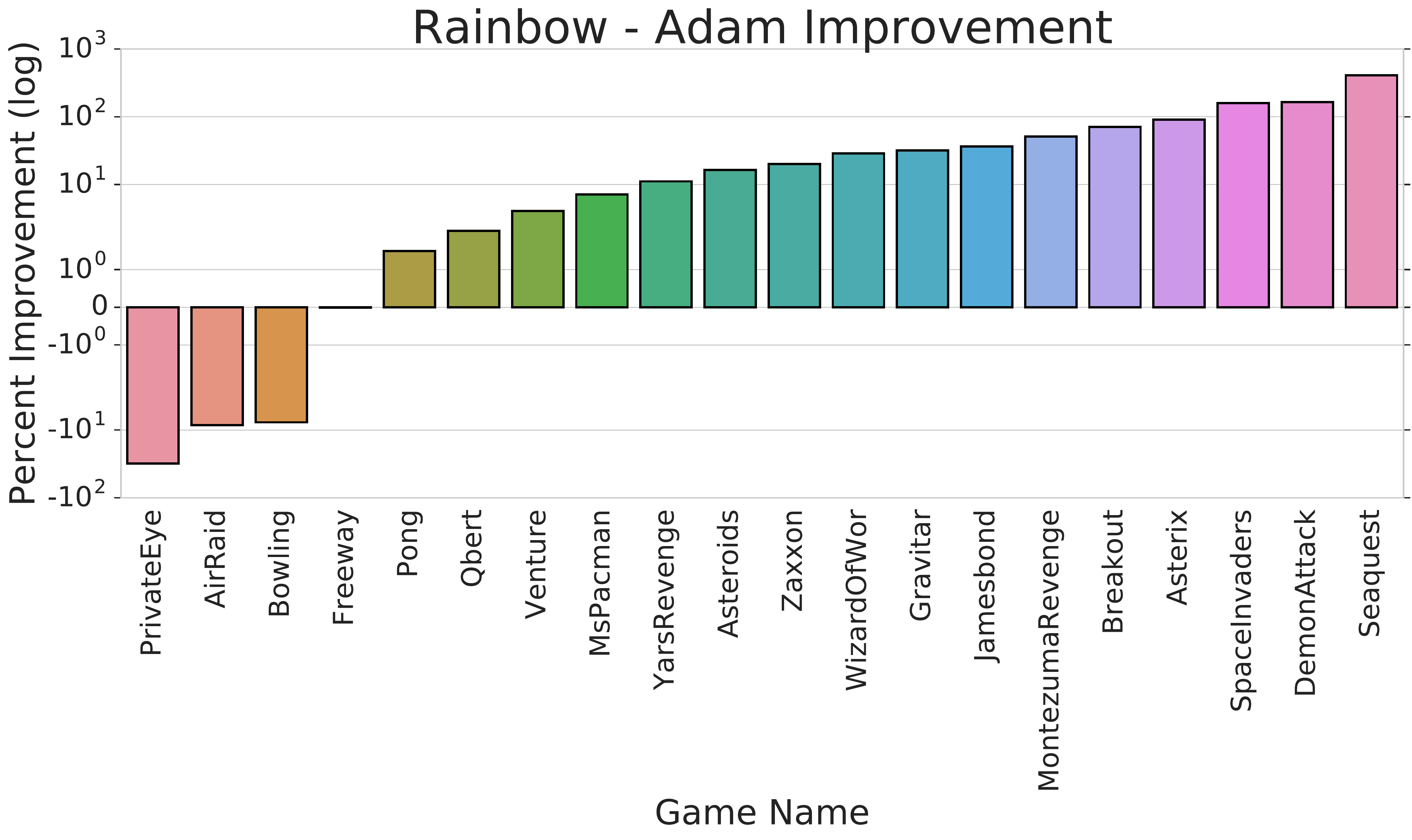}}
    \hfill
    \subfigure[The performance difference for a Rainbow agent without C51 optimizer when increasing the replay buffer size from 1M to 10M.  Median performance change of +26.6\%.]{\label{fig: rainbow_c51_10M_log}    \includegraphics[width=0.4\textwidth]{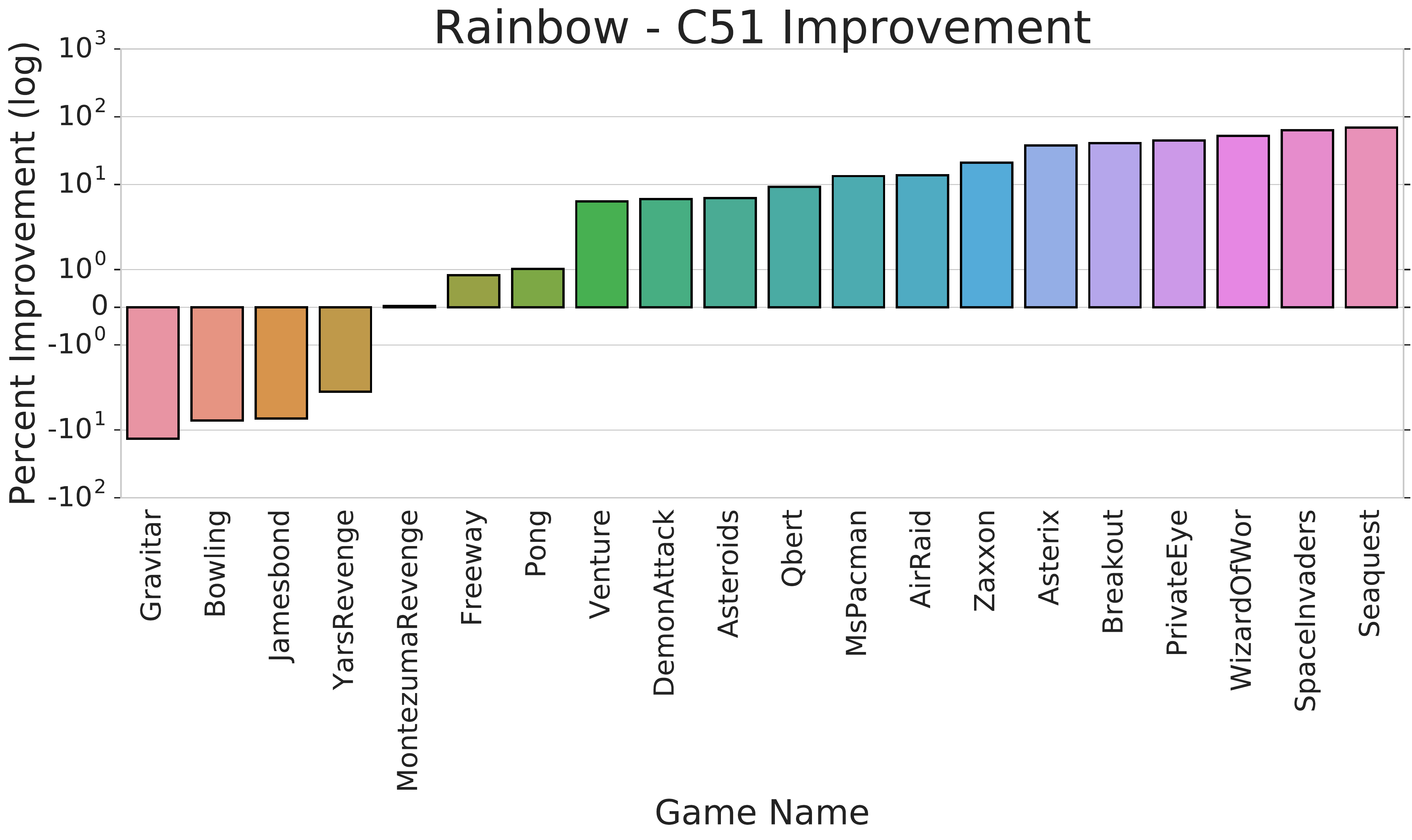}}
    \hfill
    \null
    
    \caption{Rainbow ablation results at a per-game level.}
    \label{fig: rainbow_ablation}
\end{figure}

We provide game-level granularity on the performance of each ablated Rainbow agent in Figure \ref{fig: rainbow_ablation}.

\section{Error analysis for rainbow grid}
\begin{figure}[!t]
    \centering
    \includegraphics[width=0.47\textwidth]{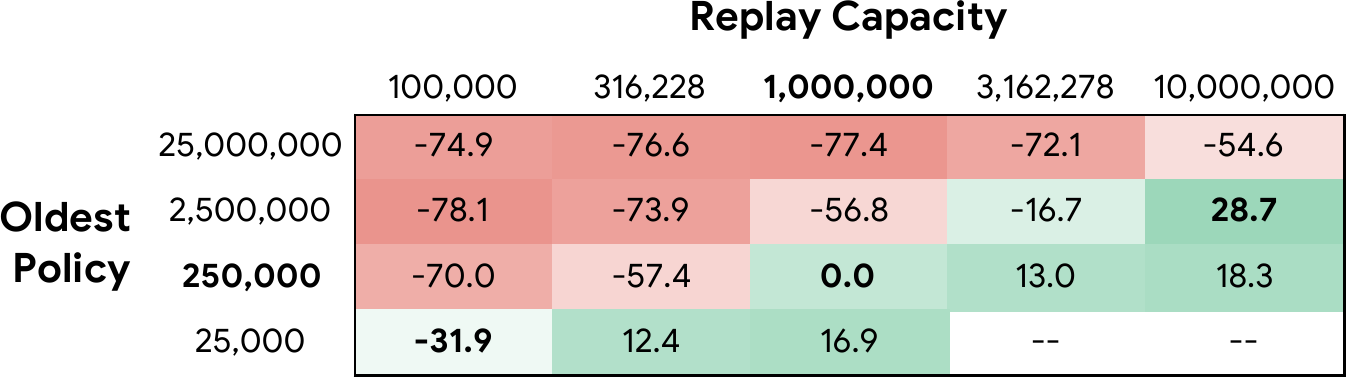}
    \caption{\textbf{Performance consistently improves with increased replay capacity and generally improves with reducing the age of the oldest policy.} We reproduce the median percentage improvement over the Rainbow baseline when varying the replay capacity and age of the oldest policy in Rainbow on a 14 game subset of Atari.}
    \label{fig: reproduced_table}
\end{figure}

\begin{figure}[!t]
    \centering
    \null
    \hfill
    \includegraphics[width=0.47\textwidth]{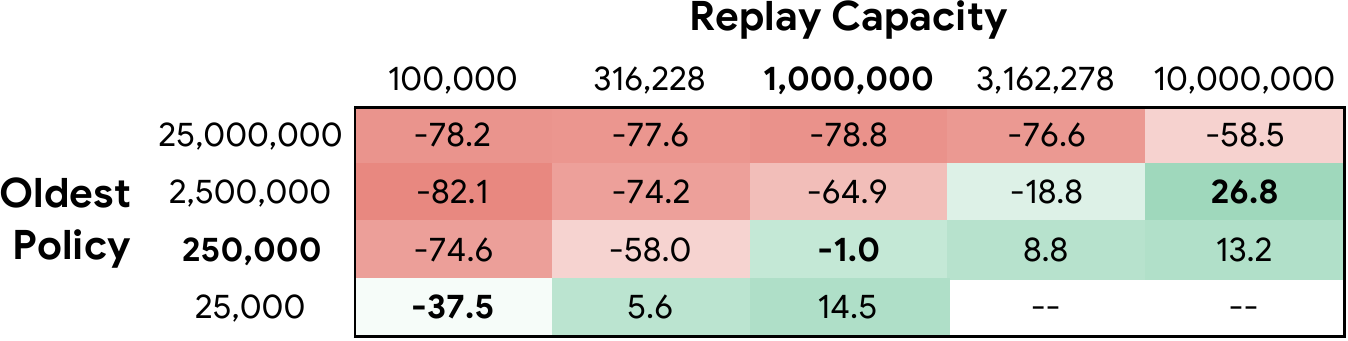}
    \hfill%
    \includegraphics[width=0.47\textwidth]{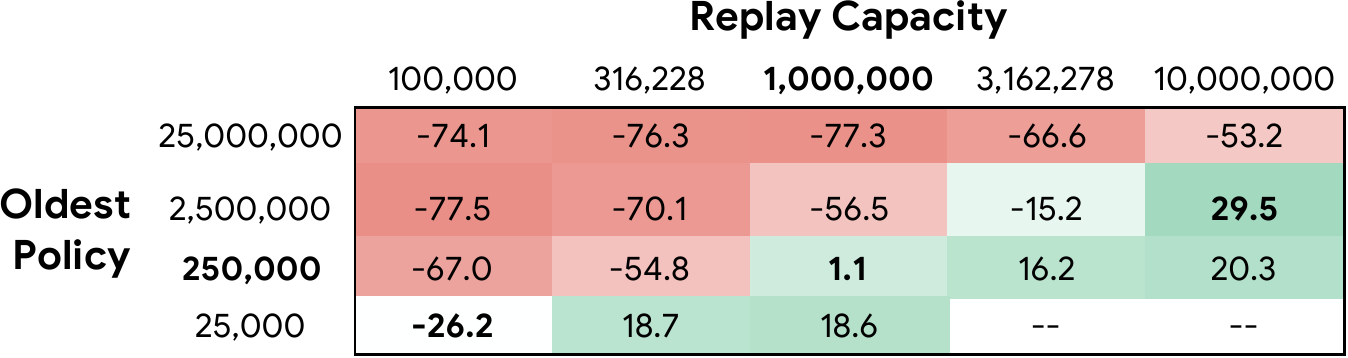}
    \hfill
    \null
    \caption{25\% (left) and 75\% (right) percentile performance improvement over the Rainbow baseline when the replay capacity and age of the oldest policy in Rainbow on a 14 game subset of Atari.}
    \label{fig: quantile_improvements}
\end{figure}

We provide an error analysis for each of the elements in Figure \ref{fig: rainbow_grid} (reproduced here as Figure \ref{fig: reproduced_table}) by providing the 25\% and 75\% percentile improvements for each combination of replay capacity and oldest policy. These results are given in Figure \ref{fig: quantile_improvements}.

\begin{figure}[!t]
    \centering
    \null
    \hfill
    \includegraphics[width=0.47\textwidth]{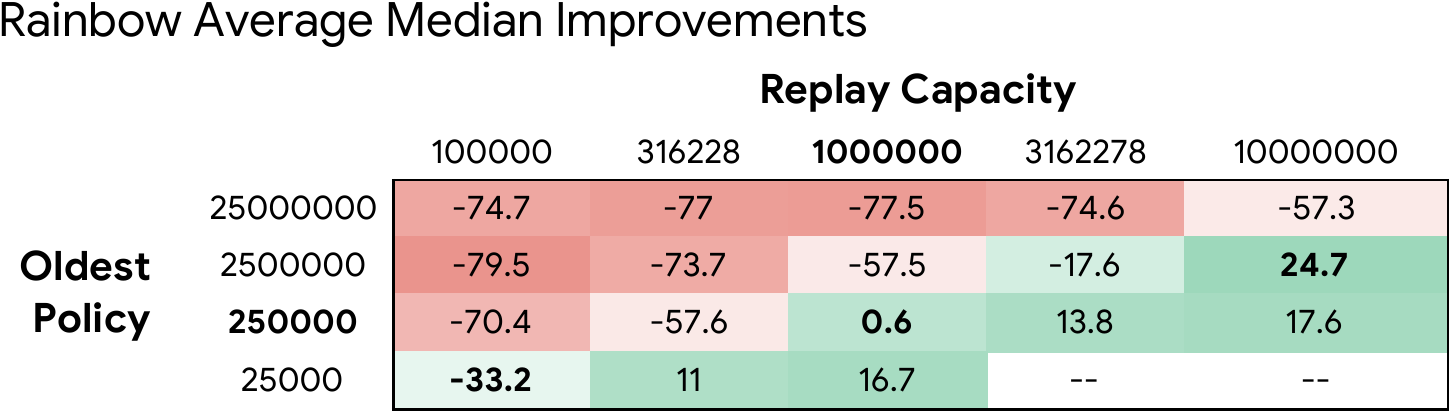}
    \hfill%
    \includegraphics[width=0.47\textwidth]{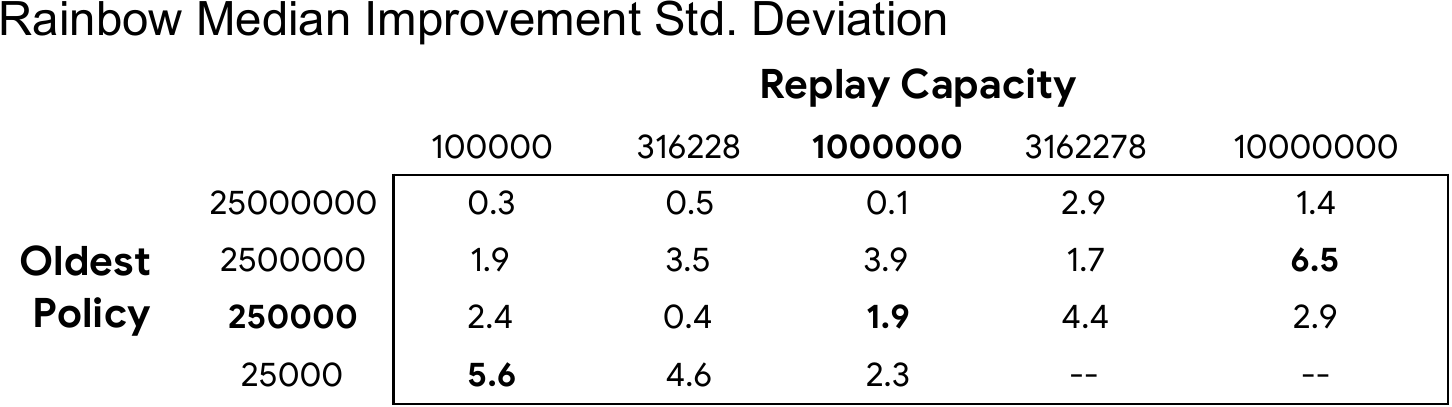}
    \hfill
    \null
    \caption{\textbf{Bootstrap estimate of variance for each cell.} For each cell, rather than using the same 3 seeds for the baseline and the same 3 seeds for each agent of each cell, we consider sampling seeds with replacement. The left grid shows the mean median improvement and the right grid shows the standard deviation of the median improvement.}
    \label{fig: rainbow_boostrap}
\end{figure}

We present an alternative view of the data using a bootstrap estimation technique.
Instead of fixing the seeds for both the baseline agent and our new agent at each cell, we sample, with replacement, the seeds.
We carry out this procedure repeatedly and report the mean and standard deviations in Figure \ref{fig: rainbow_boostrap}.

\begin{figure}[!t]
    \centering
    \includegraphics[width=0.47\textwidth]{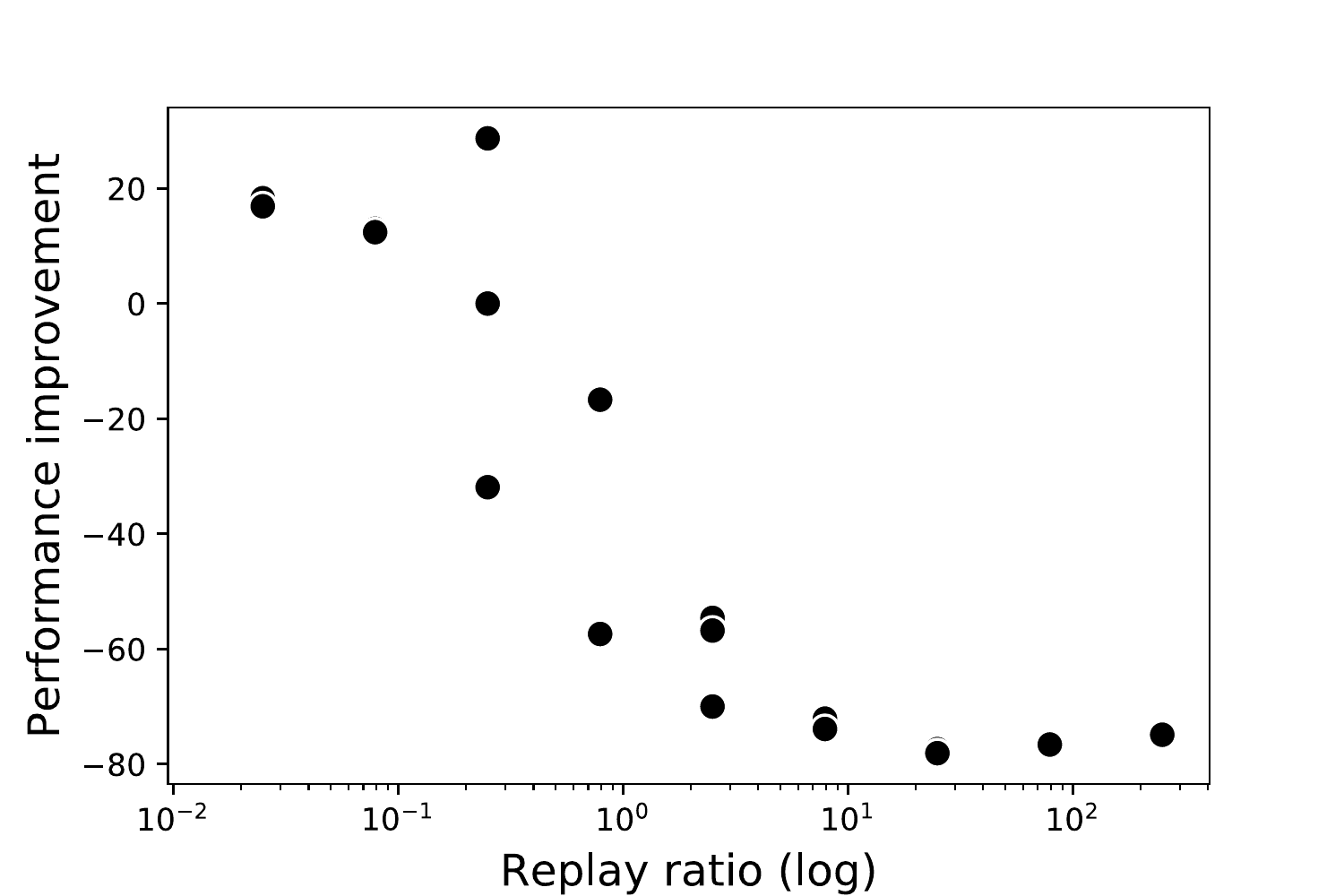}
    \caption{\textbf{Performance improvements increase as replay ratio drops.} We plot the results of Figure~\ref{fig: rainbow_grid} with respect to the replay ratio, which shows the general trend.}
    \label{fig: perf_vs_ratio}
\end{figure}

\begin{figure}

    \null
    \hfill
    \subfigure[DQN does not show benefits of replay larger than 1M, unless older policies are used]{\label{fig: lineplot1}\includegraphics[width=.75\textwidth]{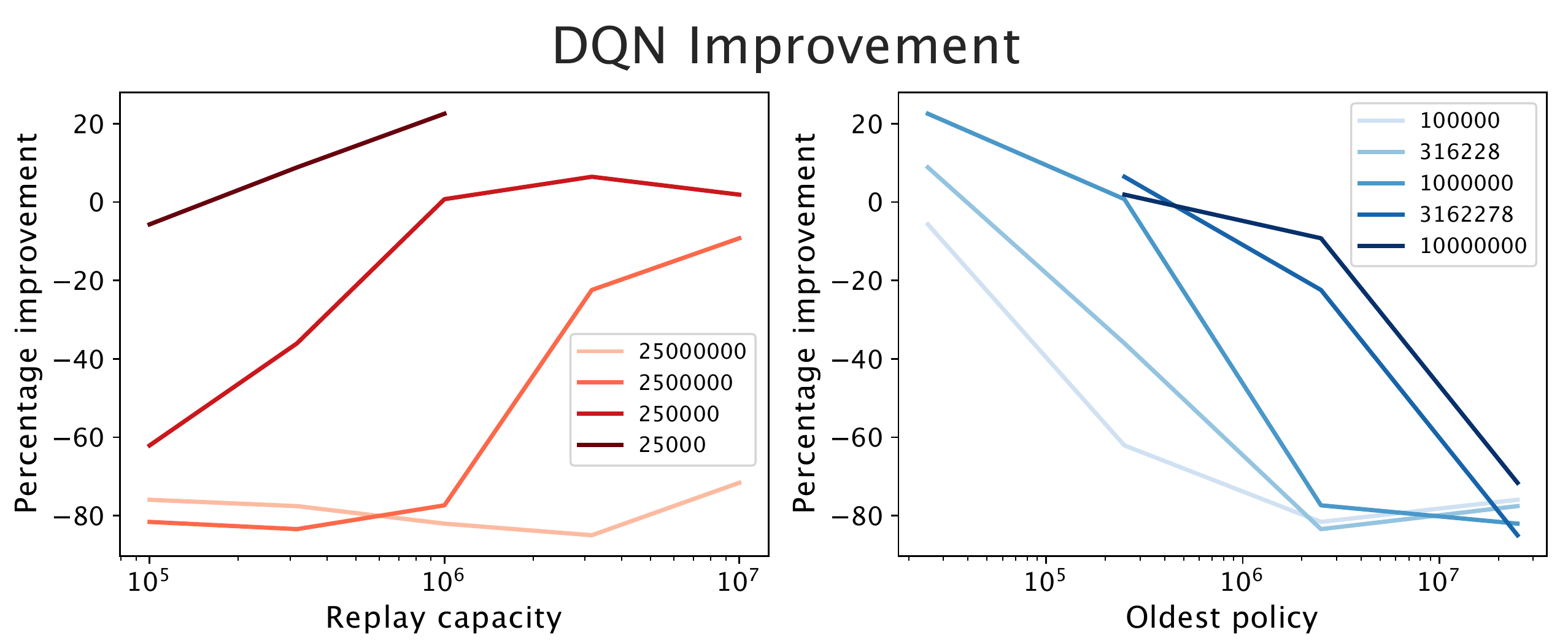}}
    \hfill
    \null

    \centering
    \null
    \hfill
    \subfigure[3-step DQN continues to improve with larger replay capacity]{\label{fig: lineplot2}\includegraphics[width=.75\textwidth]{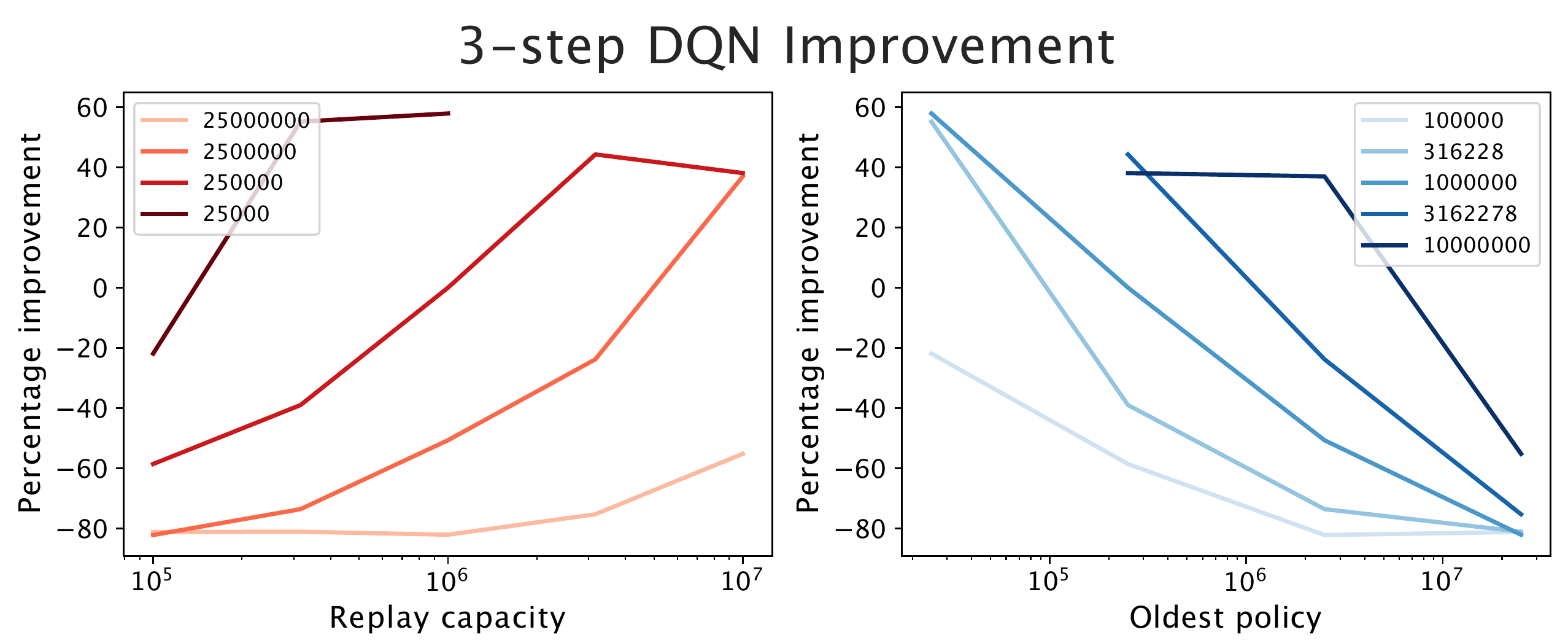}}
    \hfill
    \null
    
    \null
    \hfill
    \subfigure[Rainbow continues to improve with larger replay capacity]{\label{fig: lineplot3}\includegraphics[width=.75\textwidth]{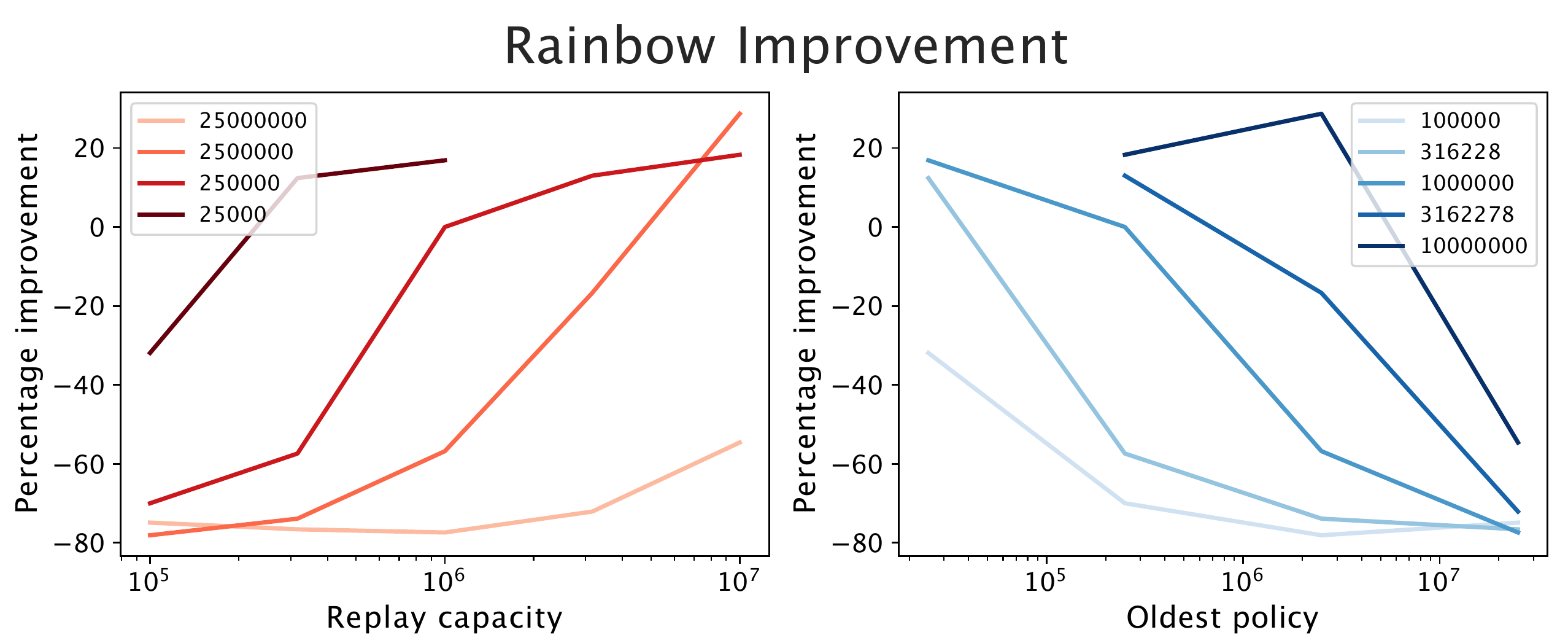}}
    \hfill
    \null
    
    \caption{Performance improves with increased replay capacity and reduced oldest policy age.}
    \label{fig: dqn_and_rainbow_lines}
\end{figure}

\section{Replay buffer size}
\begin{figure}

    \null
    \hfill
    \subfigure[1M to 100K buffer.  Median improvement: -24.6\%.]{\label{fig: rainbow_100k}\includegraphics[width=0.45\textwidth]{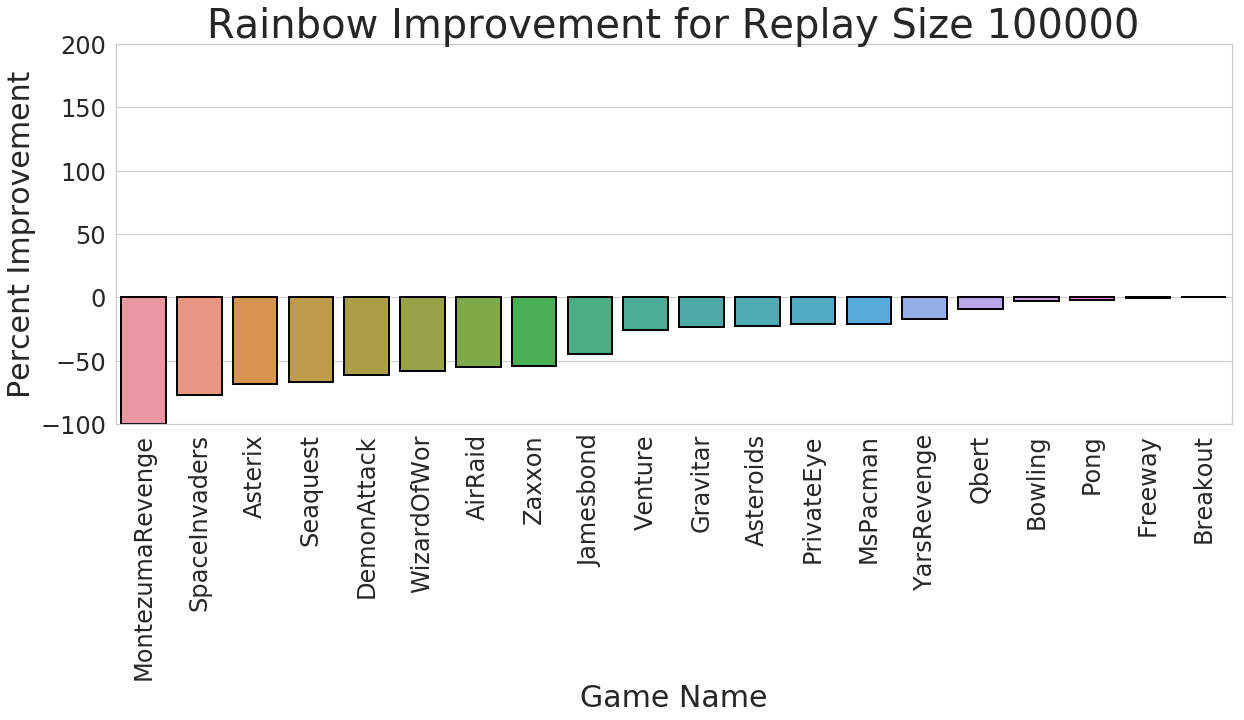}}
    \hfill
    \null

    \centering
    \null
    \hfill
    \subfigure[1M to 316K buffer.  Median improvement: -11.8\%.]{\label{fig: rainbow_316k}\includegraphics[width=0.45\textwidth]{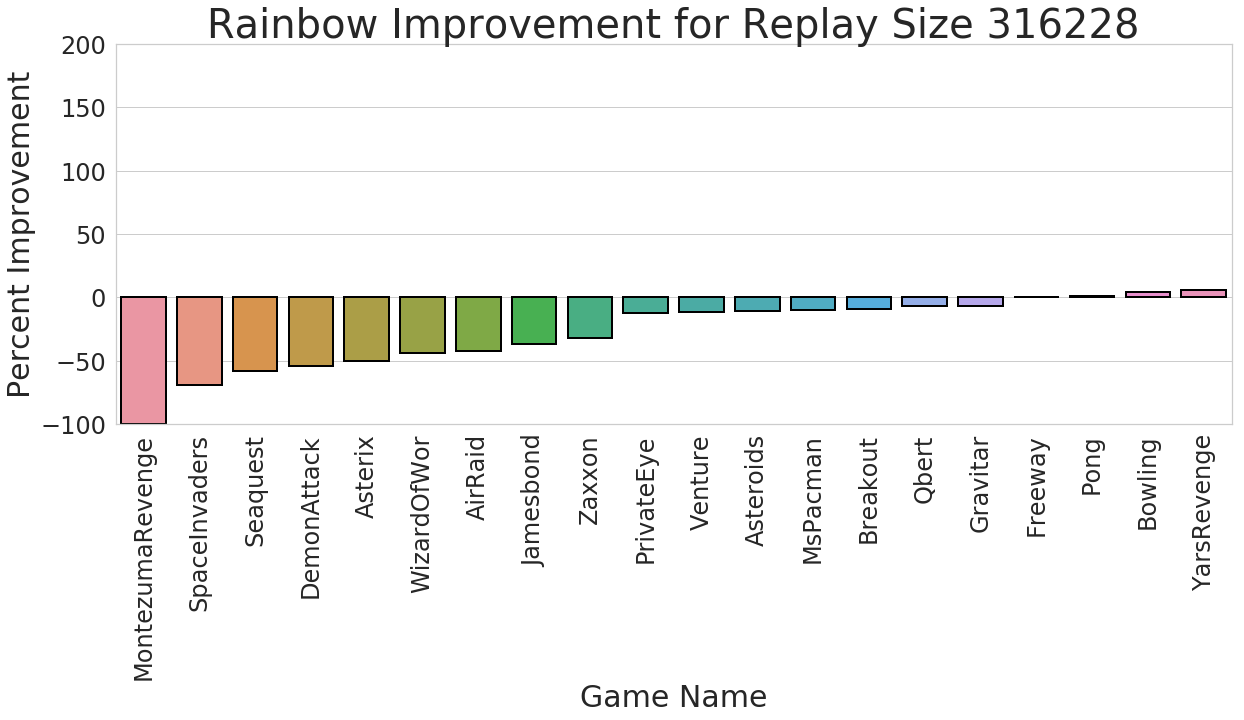}}
    \hfill
    \subfigure[1M to 1M buffer.  Median improvement: 1.6\%.]{\label{fig: rainbow_1M}    \includegraphics[width=0.45\textwidth]{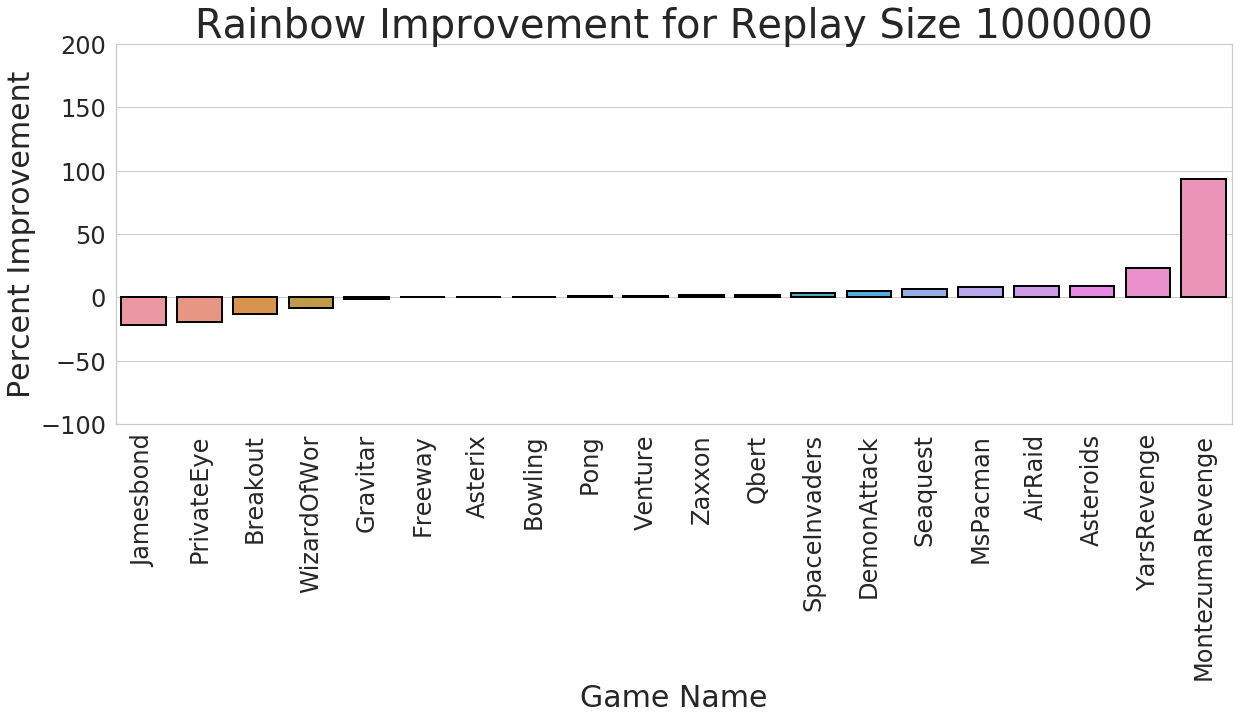}}
    \hfill
    \null
    
    \null
    \hfill
    \subfigure[1M to 3M buffer.  Median improvement: 30.2\%.]{\label{fig: rainbow_3M}\includegraphics[width=0.45\textwidth]{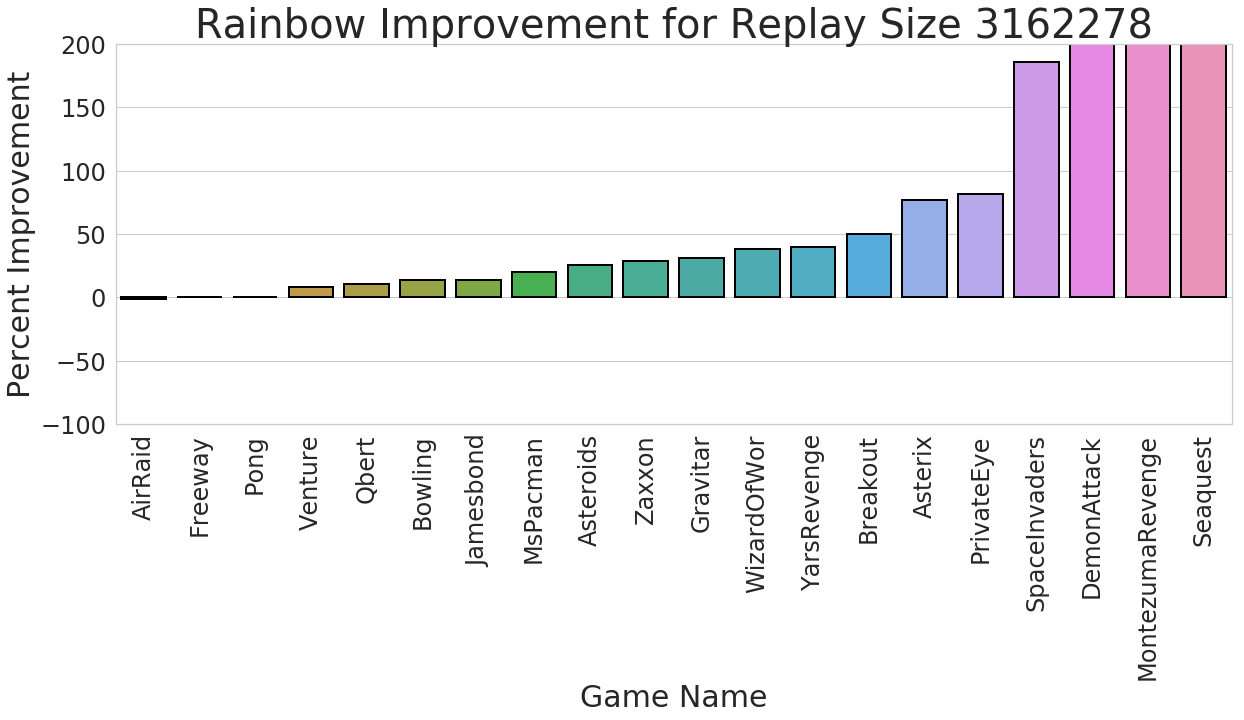}}
    \hfill
    \subfigure[1M to 10M buffer.  Median improvement: 41.0\%.]{\label{fig: rainbow_10M}    \includegraphics[width=0.45\textwidth]{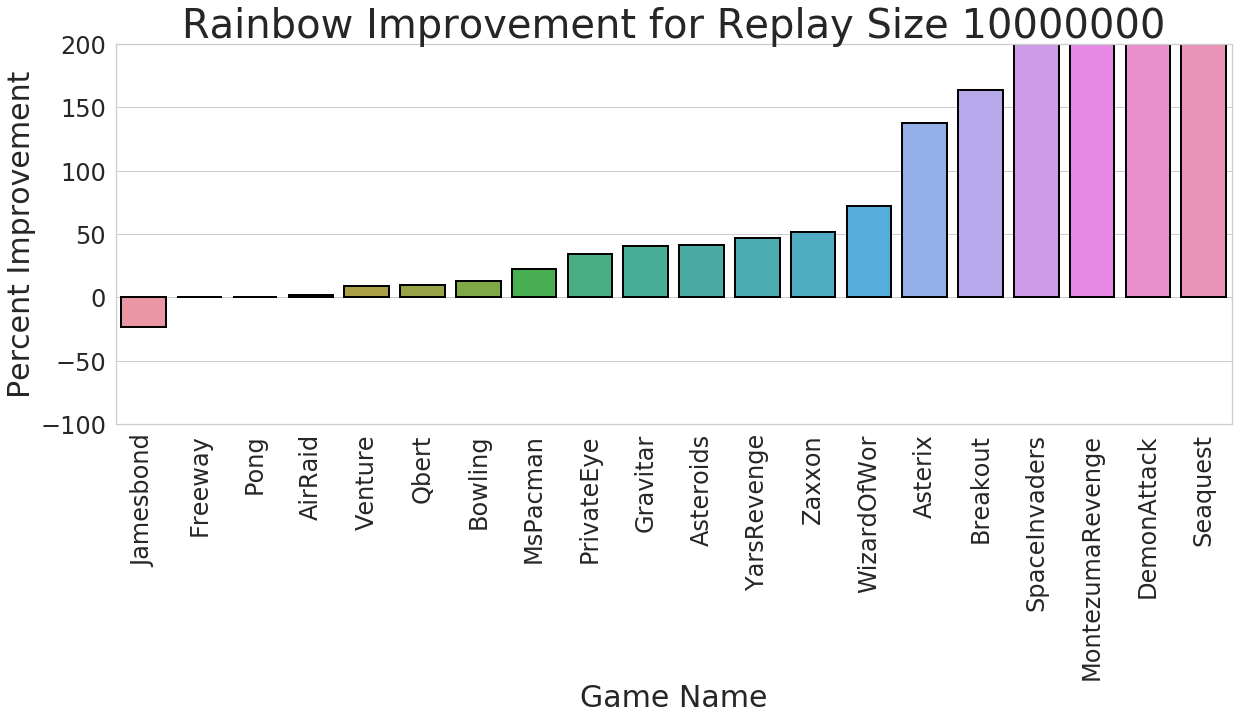}}
    \hfill
    \null
    
    \caption{Rainbow replay buffer effects at a per-game level.}
    \label{fig: rainbow_buffer}
\end{figure}

We provide a different perspective on the data from Figure \ref{fig: rainbow_grid} in Figure \ref{fig: perf_vs_ratio}, illustrating a general relationship between replay ratio and performance improvement. 
We provide game-level granularity on the performance of Rainbow with varying buffer sizes in Figure \ref{fig: rainbow_buffer}. In Figure~\ref{fig: dqn_and_rainbow_lines} we also gives results for varying replay buffer size and age of oldest policy for DQN, $3$-step DQN, and Rainbow.

\section{Batch RL learning curves}

\begin{figure*}[!ht]
    \centering
    \includegraphics[width=0.95\textwidth]{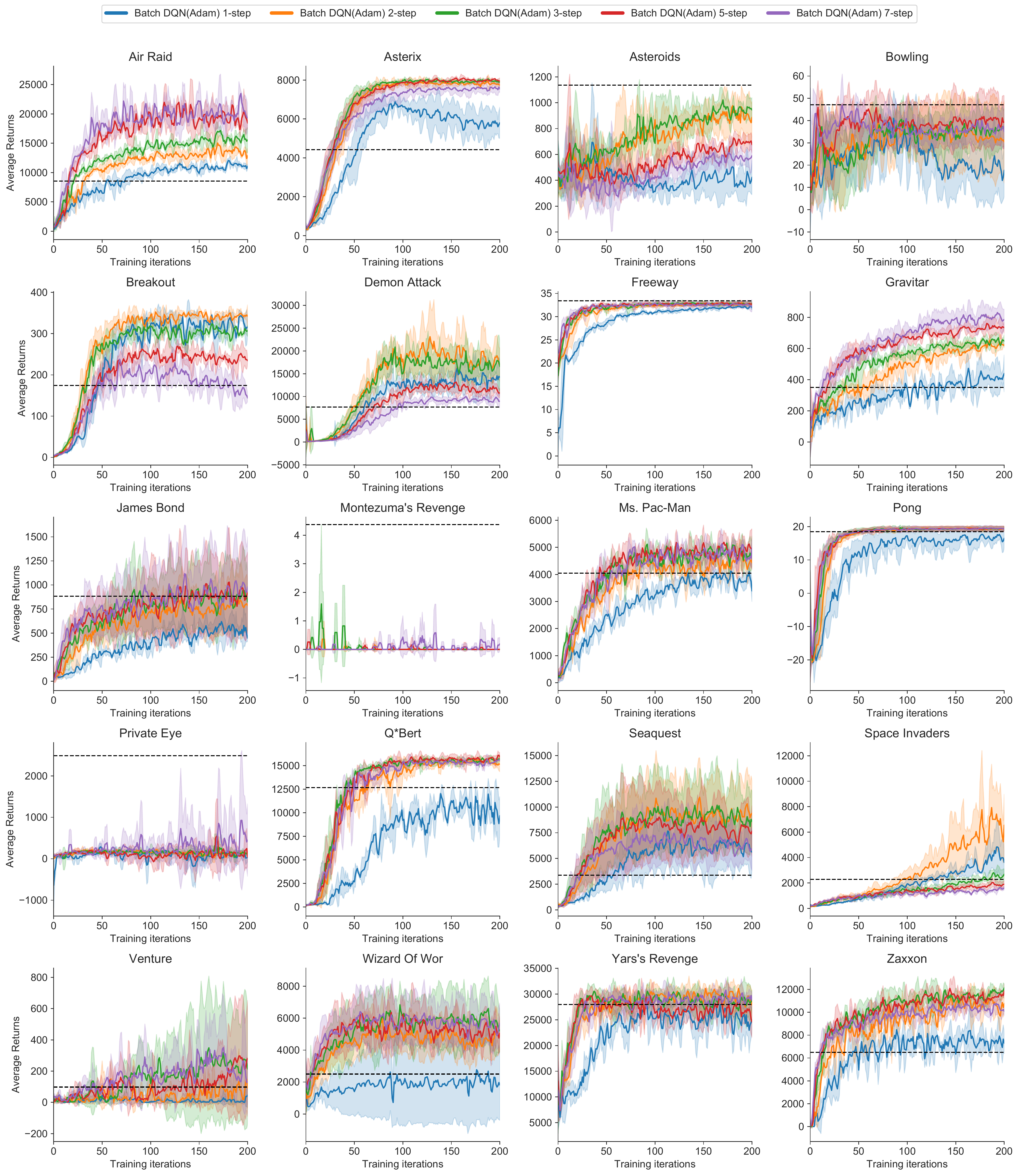}
    \caption{Average evaluation scores across 20 Atari 2600 games of batch DQN (Adam) agent with different $n$-step horizons trained offline 
    using the DQN replay dataset~\citep{agarwal2019striving}. The horizontal line shows the evaluation performance of a fully-trained online DQN agent. 
    The scores are averaged over 3 runs (shown as traces) and smoothed over a sliding window of 3 iterations and error bands show standard deviation.}
    \label{fig: dqn_batch}
\end{figure*}

\begin{figure*}[!ht]
    \centering
    \includegraphics[width=0.95\textwidth]{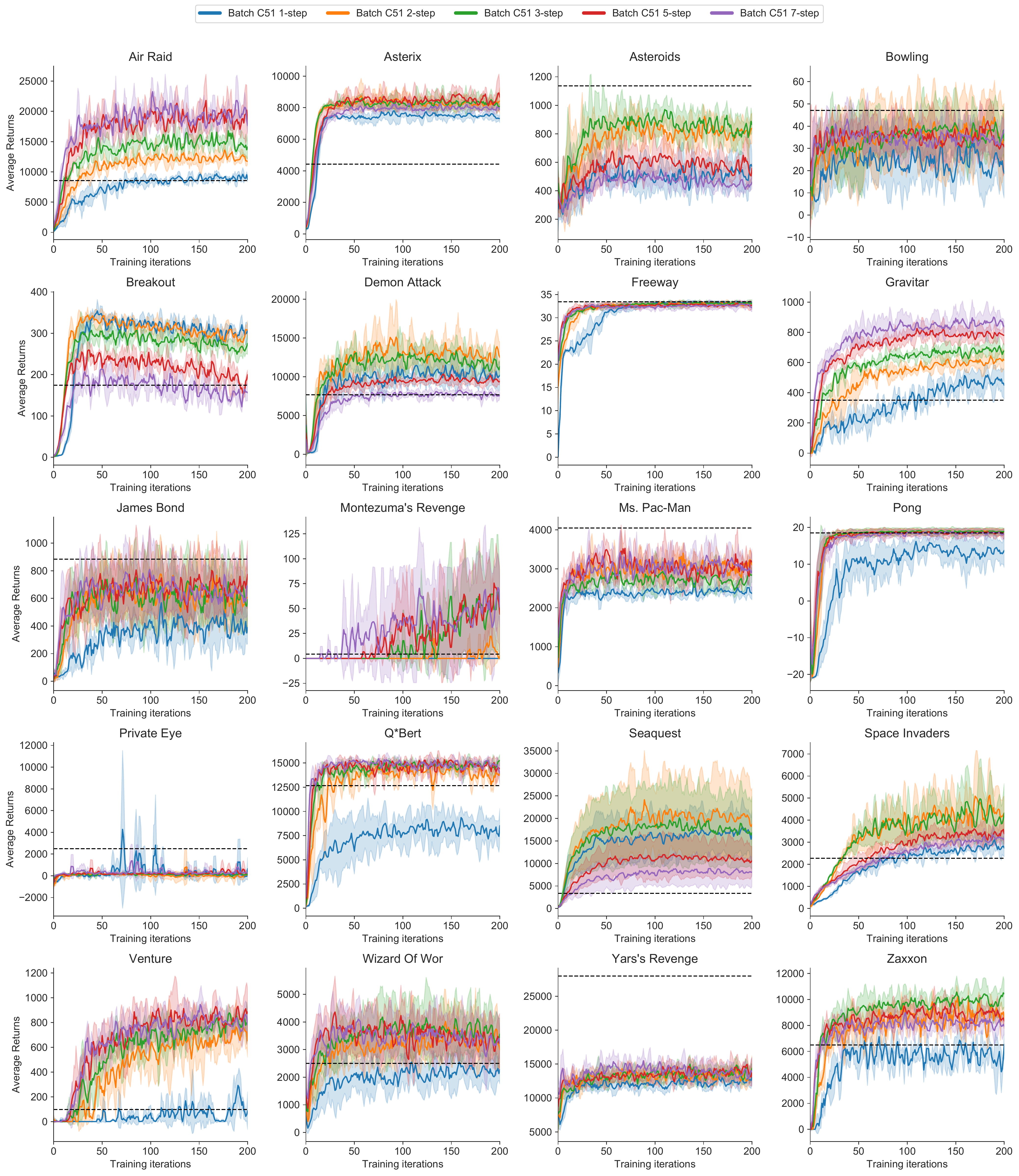}
    \caption{Average evaluation scores across 20 Atari 2600 games of a batch C51 agent with different $n$-step horizons trained offline 
    using the DQN replay dataset~\citep{agarwal2019striving}. The horizontal line shows the evaluation performance of a fully-trained online DQN agent. 
    The scores are averaged over 3 runs (shown as traces) and smoothed over a sliding window of 3 iterations and error bands show standard deviation.}
    \label{fig: c51_batch}
\end{figure*}

In Figures \ref{fig: dqn_batch} and \ref{fig: c51_batch}, we provide learning curves for the batch RL agents described in Section \ref{sec:batch}.

\end{document}